# An algorithm for a fairer and better voting system

### Gabriel-Claudiu Grama

## I. ABSTRACT

The major finding, of this article, is an ensemble method, but more exactly, a novel, better ranked voting system (and other variations of it), that aims to solve the problem of finding the best candidate to represent the voters. We have the source code [1] on GitHub, for making realistic simulations of elections, based on artificial intelligence for comparing different variations of the algorithm, and other already known algorithms.

We have convincing evidence that our algorithm is better than Instant-Runoff Voting, Preferential Block Voting, Single Transferable Vote, and First Past The Post (if certain, natural conditions are met, to support the wisdom of the crowds). By also comparing with the best voter, we demonstrated the wisdom of the crowds, suggesting that democracy (distributed system) is a better option than dictatorship (centralized system), if those certain, natural conditions are met.

Voting systems are not restricted to voting, they are ensemble methods for artificial intelligence, but the context of this article is natural intelligence. It is important to find a system that is fair (e.g. freedom of expression on the ballot exists), especially when the outcome of the voting system has social impact: some voting systems have the unfair inevitability to trend (over time) towards the same two major candidates (Duverger's law).

## II. INTRODUCTION

YouTube video available [2], and also the table of contents at the last page. In this article we present, and assess, a novel voting system (an algorithm that solves the problem of finding the best candidate to represent the voters). The assessments are done with the help of multiple independent simulations of independent elections, based on artificial intelligence.

"It is possible that the many, though not individually good men, yet when they come together may be better, not individually but collectively, than those who are so" (Aristotle, Politics, 1281a-b [3])

"Whenever you can, count" (Francis Galton [4])

Even when it comes to voting, we have have strong and objective arguments to support elitism, but we must not underestimate "the wisdom of the crowds": from a certain perspective, each individual voter is like a neuron (or a group of neurons) having it's own contribution, in a neural network, therefore, under the right conditions, a large, unbiased, heterogeneous group of voters may have very valuable emergent properties.

"The wisdom of the crowds" was observed at a 1906 country fair in Plymouth by statistician and polymath Francis Galton [5]. Economist Jack Treynor conducted the "jelly beans in a jar" experiment [6]. However,

Aristotle is considered to be the first, to write about "the wisdom of the crowds", in his work of political philosophy, titled "Politics" [3].

Let's consider Arrow's Impossibility Theorem, and the criteria which scholars use to rate a voting system: Majority criterion, Mutual majority criterion, Condorcet winner criterion, Condorcet loser criterion, Independence of irrelevant alternatives criterion, Independence of clones criterion. For a better awareness of the current state of the art in the field, consider investigating ensemble methods, honorable mentions of voting systems such as Instant-Runoff Voting, and also, the following articles might be of interest: [7] [8] [9] (note that the unaffiliated author did not read these three closed-access articles).

## III. PROPOSED METHODOLOGY

### A. The Basics

In our solution, the voter is supposed to express his desire (or educated guess), in an explicit manner, as illustrated in the next image:

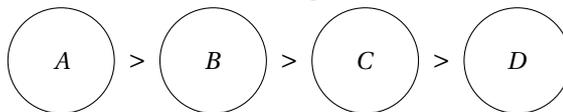

In the previous image, we have the representation of a ballot. A,B,C,D are candidates, and A is the most favorite option, B is the second favorite option, C is the third, and D is the fourth.

If the ballot does not contain duplicates of candidates, we call it a "valid ballot". If a ballot contains all the candidates, we call it a "complete ballot".

In this example, B could also be $\varnothing$ (null candidate), thus this means that C and D are undesirable.

It is fairer to add the $\varnothing$ candidate, as a mean to express protest. It is also fairer to add the "I don't know" candidate as a mean to express abstinence (while these votes are being ignored). It is fairer to add these candidates, first in the ballot, because it can teach or remind voters that these options exist. It might also encourage people to go to vote, and even if they choose the "I don't know" candidate, the voters are showing that their class is voting thus they are relevant and not negligible by candidates (if votes are secret).

### B. The Algorithm

We will begin by describing the basic version of the algorithm (more at III-D).

To decide the winner, we must first pick an arbitrary $\alpha$, $0 \le \alpha \le 1$, before the voting process starts (for ethical reasons). For example, if $\alpha = 0.5$ this means that the winner must pass the threshold score of $(100 \cdot \alpha)\%$. We want $\alpha = 0.5$ when we want the candidate winner to represent more than 50% of the voters.

As shown in the next tables, to make use of our counted votes, we need to find a score for each candidate, from the most preferred option to the least preferred option, in order, stage by stage. If nobody has won in the current stage, we proceed to the next stage by adding the votes of the current stage in to the votes of the next unprocessed stage. In other words, the voters who had a higher preference, join forces with those of lower preference until somebody wins. If they are multiple candidates that pass the threshold, the one with the highest score wins.

We do this addition because we don't want higher preferences to be lost, they matter, there is information to be used. We proceed to the next stage because, based on $\alpha$, we didn't find a winner.

We want $\varnothing$ to be treated as a candidate, it plays a very important role in the algorithm.

The first step is to count the ballots, to construct the next table:

| Vote Counts | | | | | |
|---|---|---|---|---|---|
| | A | B | C | D | ... |
| Preference 1 | $a_1$ | $b_1$ | $c_1$ | $d_1$ | ... |
| Preference 2 | $a_2$ | $b_2$ | $c_2$ | $d_2$ | ... |
| Preference 3 | $a_3$ | $b_3$ | $c_3$ | $d_3$ | ... |
| Preference 4 | $a_4$ | $b_4$ | $c_4$ | $d_4$ | ... |
| ... | ... | ... | ... | ... | ... |

Next, we use the function $f_1$ to obtain the next Table (Processed Vote Counts):

$$f_{1(X,i)} = \sum_{j \le i} x_j = x_1 + x_2 + x_3 + ... + x_i = f_{1(X,i-1)} + x_i$$

$f_{1(X,i)}$ = processed vote count for candidate $X$ at stage $i$; $x_j$ = vote count for candidate $X$ as preference ; $i, j \in \mathbb{N}^*$

| Processed Vote Counts | | | | | |
|---|---|---|---|---|---|
| | A | B | C | D | ... |
| Stage1 | $a_1$ | $b_1$ | $c_1$ | $d_1$ | ... |
| Stage2 | $a_1 + a_2$ | $b_1 + b_2$ | $c_1 + c_2$ | $d_1 + d_2$ | ... |
| Stage3 | $a_1 + a_2 + a_3$ | $b_1 + b_2 + b_3$ | $c_1 + c_2 + c_3$ | $d_1 + d_2 + d_3$ | ... |
| Stage4 | $a_1 + a_2 + a_3 + a_4$ | $b_1 + b_2 + b_3 + b_4$ | $c_1 + c_2 + c_3 + c_4$ | $d_1 + d_2 + d_3 + d_4$ | ... |
| ... | ... | ... | ... | ... | ... |

After processing the votes, we can find the score of each candidate for each stage (we only use the table with the Processed Vote Counts). We use the $f_2$ function, to obtain the next table (Score of Candidates):

$$f_{2(X,i)} = \frac{100 \cdot f_{1(X,i)}}{i \cdot n}$$

$f_{2(X,i)}$ = processed vote count of candidate $X$ at stage $i$; $n$ = number of voters; $i, n \in \mathbb{N}^*$

We multiply by 100 and divide by $i \cdot n$ because we need to know, in percentages, how many voters voted for the associated candidate, up to the specific stage i, and we assume that every voter submitted a complete ballot. Sorting the columns helps when two candidates have equal best score on the same stage.

| Score of Candidates | | | | | |
|---|---|---|---|---|---|
| | A | B | C | D | ... |
| Stage1 | $\frac{100 \cdot f_{1(A,1)}}{1 \cdot n}$ | $\frac{100 \cdot f_{1(B,1)}}{1 \cdot n}$ | $\frac{100 \cdot f_{1(C,1)}}{1 \cdot n}$ | $\frac{100 \cdot f_{1(D,1)}}{1 \cdot n}$ | ... |
| Stage2 | $\frac{100 \cdot f_{1(A,2)}}{2 \cdot n}$ | $\frac{100 \cdot f_{1(B,2)}}{2 \cdot n}$ | $\frac{100 \cdot f_{1(C,2)}}{2 \cdot n}$ | $\frac{100 \cdot f_{1(D,2)}}{2 \cdot n}$ | ... |
| Stage3 | $\frac{100 \cdot f_{1(A,3)}}{3 \cdot n}$ | $\frac{100 \cdot f_{1(B,3)}}{3 \cdot n}$ | $\frac{100 \cdot f_{1(C,3)}}{3 \cdot n}$ | $\frac{100 \cdot f_{1(D,3)}}{3 \cdot n}$ | ... |
| Stage4 | $\frac{100 \cdot f_{1(A,4)}}{4 \cdot n}$ | $\frac{100 \cdot f_{1(B,4)}}{4 \cdot n}$ | $\frac{100 \cdot f_{1(C,4)}}{4 \cdot n}$ | $\frac{100 \cdot f_{1(D,4)}}{4 \cdot n}$ | ... |
| ... | ... | ... | ... | ... | ... |

## C. Concrete example

In this example, we illustrate one of the important challenges that this approach solves. In the next table, we observe that everyone has his best favorite, stamps being distributed evenly between candidates, except, candidate X is nobodies favorite, but X is everyone's second favorite. If $\alpha$=0.5 then the winner is X, because nobody had enough votes as first preference to win.

For simplicity let's assume they are 100 voters. We will also consider $\varnothing$ to be the null candidate, to make it clear that everyone has only 2 preferences, but would choose between a bad and a worse candidate, if none of their preferred candidates are strong enough to win (their first preference can still win at later stages). We remind that voters can specify 6 preferences because that's also the number of candidates ($\varnothing$ is a candidate).

The three tables for the concrete example:

| Vote Counts | | | | | | |
|---|---|---|---|---|---|---|
| | A | B | C | D | X | $\varnothing$ |
| Preference1 | 25 | 25 | 25 | 25 | 0 | 0 |
| Preference2 | 0 | 0 | 0 | 0 | 100 | 0 |
| Preference3 | 0 | 0 | 0 | 0 | 0 | 100 |
| Preference4 | 25 | 25 | 25 | 25 | 0 | 0 |
| Preference5 | 25 | 25 | 25 | 25 | 0 | 0 |
| Preference6 | 25 | 25 | 25 | 25 | 0 | 0 |

| Processed Vote Counts | | | | | | |
|---|---|---|---|---|---|---|
| | A | B | C | D | X | $\varnothing$ |
| Stage 1 | 25 | 25 | 25 | 25 | 0 | 0 |
| Stage 2 | 25 | 25 | 25 | 25 | 100 | 0 |
| Stage 3 | 25 | 25 | 25 | 25 | 100 | 100 |
| Stage 4 | 50 | 50 | 50 | 50 | 100 | 100 |
| Stage 5 | 75 | 75 | 75 | 75 | 100 | 100 |
| Stage 6 | 100 | 100 | 100 | 100 | 100 | 100 |

| Score of Candidates | | | | | | |
|---|---|---|---|---|---|---|
| | A | B | C | D | X | $\varnothing$ |
| Stage1 | 25% | 25% | 25% | 25% | 0% | 0% |
| Stage2 | 25% | 25% | 25% | 25% | 100% | 0% |
| Stage3 | 25% | 25% | 25% | 25% | 100% | 100% |
| Stage4 | 50% | 50% | 50% | 50% | 100% | 100% |
| Stage5 | 75% | 75% | 75% | 75% | 100% | 100% |
| Stage6 | 100% | 100% | 100% | 100% | 100% | 100% |

## D. A better version of the algorithm

This section, is all about deriving: we present some of the features that an algorithm variant can have.

• In one version, we select an arbitrary $\alpha = 0.5$, and keep advancing in stages until $\varnothing$ passes the threshold.

We do not proceed to the next stages, if $\varnothing$ passed the threshold $\alpha$. If none of the candidates passed the threshold, until this point, then $\varnothing$ is the winner. If they are other candidates, that passed the threshold, then the candidate with the best score wins, even if they are candidates that passed the threshold at earlier stages. A non-$\varnothing$ candidate cannot be a winner on a stage where $\varnothing$ has a better score, so in this case, we need to select a winner from previous stages, but must still pass the threshold.

In other words, as long the voters are content with the candidates, we want to keep on measuring the value of each candidate, using the "wisdom of the crowd".

If we settle with this approach, then ballots must support at least $k-1$ preferences, where $k$ is the number of candidates (including $\varnothing$).

Let's consider the following example:

| Score of Candidates | | | | | |
|---|---|---|---|---|---|
| | A | B | C | D | $\varnothing$ |
| Stage 1 | 25% | 25% | 25% | 25% | 0% |
| Stage 2 | 65% | 55% | 40% | 40% | 0% |
| Stage 3 | $w$% | $x$% | $y$% | $z$% | $q > 50\%$ |

In this example, we stopped at stage 3, because that's when $\varnothing$ passed the threshold with a value of q (assuming $\alpha = 0.5$). If $q$ is the highest score (at stage 3), then A is the winner, with a score of 65%. Otherwise, the winner will be decided based on $w, x, y, z$.

However, with this approach, we risk ending up choosing the winner from a stage with candidates of similar score, as a result of approaching the last possible stage, from a lack of votes for $\varnothing$.

• As a counter measure, we may want to stop advancing in stages, once a candidate has passed a score of (for example) $\gamma = 66.66\%$ (two thirds).

• Another idea, is to have a second threshold only used for $\varnothing$, for example an arbitrary $\beta = 0.3333$ (one third). In other words, we advance in stages until $\varnothing$ passes the threshold $\beta$, we stop when we cannot tolerate more voters to be unhappy. In this case, $\beta$ dictates the maximum number of voters we allow to be unhappy.

⚠ Since we stopped at a stage where too many voters are unhappy (more than 33.33% voted for $\varnothing$), then maybe it's better to only decide the winner based on previous stages. So, in this case, the winner is candidate A regardless of scores $w, x, y, z$ from stage 3:

| Score of Candidates | | | | | |
|---|---|---|---|---|---|
| | A | B | C | D | $\varnothing$ |
| Stage 1 | 25% | 25% | 25% | 25% | 0% |
| Stage 2 | 65% | 55% | 40% | 40% | 0% |
| Stage 3 | $w$% | $x$% | $y$% | $z$% | $q > 33.33\%$ |

• We could go so far as to use Shannon's entropy: $H(X) = \sum_{i=1}^{n} P(x_i) \cdot \log_2 P(x_i)$. Using Score of Candidates table, $P(x_i) = \frac{f_{2(x_i, k)}}{100 \cdot k}$. Notice how $H(X)$ is maximum at the last possible stage and that hyper-parameters, like $\gamma$, can be defined using $H(X)$, and even variance: $Var(X)$. Picking a winner on the valid stage with the lowest $H(X)$, is an option, especially if $H(X)$ is also an outlier. A single outlier $H(X)$ might not be a coincidence.

## E. About Realistic simulations

These are the properties of our realistic simulation:

• a simulation has multiple elections.
• during the entire simulation, the same voting systems and the same voters are being used (for each election).
• the qualities of voters (mean squared errors) are, more or less, normally distributed
• each voter is a trained neural network (with predetermined quality to obtain the normal distribution).
• voters have emergent properties
• the population of voters is heterogeneous, because each voter is partially, randomly, permanently blind to a number of features that the candidates have. Collectively, the population sees all the features.
• candidates have multiple relevant features

Some reasons why our realistic simulation is less realistic (as of 2021):

• our voters rely on sincerity and educated guesses, they don't lie, they don't resort to tactical voting.
• our independent voters are not being manipulated by other voters, candidates, media or anything else.

In the real world, there is an issue: people don't know everything, they don't take into account all the things that need to be taken, when they make a decision, such as how to vote. This issue is being compensated with the fact that the population is heterogeneous, however, if the population is manipulated, and not able to make independent, educated guesses, that compensation is less impactful. To know and understand more about the necessities of the population, James Surowiecki wrote the book "The Wisdom Of Crowds" [10], from where Oinas-Kukkonen captures the wisdom of crowds approach with eight conjectures.

Candidates are based on our synthetic data set. Each candidate has 11 features. 10 of the features are randomly generated numbers from $[5, 10)$. The 11th feature is the weighted sum of those features. The weights are generated randomly from $[-10, 10)$. We made the synthetic data set in such a way that all features matter. We decided that the higher the 11th feature is, the better the candidate. We generated 3000 such candidates, and 30% were used only for testing purposes.

We decided that the null candidate has a 11th feature equal to the median of all 11th scores and that all of our artificial voters agree to this.

The mean and standard deviation of the aimed normal distribution, are picked based on other simulations with 50 to 200 voters, where the training is not restricted, to see how well can a voter do with a given blindness. We also checked, during the training, what the lowest quality of a voter we can expect to get, with just one epoch of training. based on the two extremes we picked a mean that is, more or less, three standard deviations away from the two extremes (exceptions may exist). In the context of this paragraph, the numbers were chosen by eye.

Ballots are generated based on the predicted 11th feature, by the voters (regression problem). Since we decided that the higher the 11th feature is, all we had to do is sort the predicted values and associated the rankings, accordingly, to generate the ballots.

We assessed a vast number of variations of our novel algorithm, and also already known algorithms (Instant-Runoff Voting, Preferential Block Voting, Single Transferable Vote, First Past The Post), for comparison. We also compared all of these algorithms, against the best voter (unit), and the whole group of voters (group as a unit; based on mean and median). In other words we also tested the quality of the best voter's ballots (assessing dictatorship), and the ballots of the crowd as a unit. In other words, when we created just 1 ballot from the whole crowd, we used the mean and median over the predictions from all voters (this is only for comparison).

We used the python package pyrankvote for Instant-Runoff Voting, Preferential Block Voting, and Single Transferable Vote.

The metrics that we used to asses each voting system, are the following:

- meanWinnerRank: the average true rank of the winners. This is, by far, the most relevant metric, and the perfect score is 1.
- rateTrueWinners: the rate at which the winner is the best candidate.
- rateWinner<NULL: the rate at which the winner is worse than the null candidate.

### F. Realistic Simulation Results (Evidence for quality)

Each following simulation output (which is a dataframe), is preceded by coresponding hyper-parameters, and followed by validation mean squared errors, if the case (which are so unexpected, that we lack an explanation).

================================ START OF SIMULATION 1 ================================

```
numCandiates  :  20
numVoters  :  500
numElections  :  500
columnBlindness  :  9
epochs  :  133
trainableLayerCount  :  1
crowdBuildMethod  :  {'name': 'standardDistribution', 'mean': 18000, 'standardDeviation': 500}
dataSetName  :  mySynthetic
predictedFeature  :  y
```

======== SIMULATION RESULTS ========

| Metrics<br><br>Algorithms | meanWinnerRank | rateTrueWinners | rateWinner<NULL |
|---|---|---|---|
| crowd−Mean | 1.464 | 0.696 | 0.000 |
| crowd−Median | 1.464 | 0.696 | 0.000 |
| MyVoteSys <$\alpha$=0.66, $\beta$=0.33, $\gamma$=____, MinEntropy> | 2.534 | 0.442 | 0.024 |
| MyVoteSys <$\alpha$=0.66, $\beta$=0.33, $\gamma$=0.80, MinEntropy> | 2.534 | 0.442 | 0.024 |
| MyVoteSys <$\alpha$=0.66, $\beta$=____, $\gamma$=0.80, MinEntropy> | 2.534 | 0.442 | 0.024 |
| MyVoteSys <$\alpha$=0.66, $\beta$=____, $\gamma$=____, MinEntropy> | 2.534 | 0.442 | 0.024 |
| MyVoteSys <$\alpha$=0.66, $\beta$=0.33, $\gamma$=0.80, First> | 2.548 | 0.444 | 0.026 |
| MyVoteSys <$\alpha$=0.50, $\beta$=0.33, $\gamma$=0.66, Last> | 2.548 | 0.444 | 0.026 |
| MyVoteSys <$\alpha$=0.50, $\beta$=____, $\gamma$=0.66, Last> | 2.548 | 0.444 | 0.026 |
| MyVoteSys <$\alpha$=0.66, $\beta$=____, $\gamma$=0.80, First> | 2.548 | 0.444 | 0.026 |
| MyVoteSys <$\alpha$=0.66, $\beta$=0.33, $\gamma$=____, First> | 2.548 | 0.444 | 0.026 |
| MyVoteSys <$\alpha$=0.66, $\beta$=____, $\gamma$=____, First> | 2.548 | 0.444 | 0.026 |
| MyVoteSys <$\alpha$=0.50, $\beta$=____, $\gamma$=0.66, MaxEntropy> | 2.564 | 0.438 | 0.026 |
| MyVoteSys <$\alpha$=0.50, $\beta$=0.33, $\gamma$=0.66, MaxEntropy> | 2.564 | 0.438 | 0.026 |
| MyVoteSys <$\alpha$=0.66, $\beta$=0.33, $\gamma$=0.80, MinVariance> | 2.604 | 0.446 | 0.026 |
| MyVoteSys <$\alpha$=0.50, $\beta$=____, $\gamma$=0.66, MinVariance> | 2.632 | 0.426 | 0.028 |
| MyVoteSys <$\alpha$=0.50, $\beta$=0.33, $\gamma$=0.66, MinVariance> | 2.632 | 0.426 | 0.028 |
| MyVoteSys <$\alpha$=0.66, $\beta$=____, $\gamma$=0.80, MinVariance> | 2.650 | 0.444 | 0.032 |
| MyVoteSys <$\alpha$=0.50, $\beta$=0.33, $\gamma$=0.80, MinVariance> | 2.664 | 0.430 | 0.028 |
| MyVoteSys <$\alpha$=0.50, $\beta$=____, $\gamma$=0.80, MinVariance> | 2.684 | 0.428 | 0.028 |
| MyVoteSys <$\alpha$=0.66, $\beta$=0.33, $\gamma$=____, MinVariance> | 2.698 | 0.436 | 0.024 |
| MyVoteSys <$\alpha$=0.50, $\beta$=0.33, $\gamma$=____, MinVariance> | 2.752 | 0.422 | 0.026 |
| MyVoteSys <$\alpha$=0.50, $\beta$=____, $\gamma$=____, MinVariance> | 2.856 | 0.414 | 0.030 |
| MyVoteSys <$\alpha$=0.50, $\beta$=0.33, $\gamma$=0.80, MaxEntropy> | 2.858 | 0.436 | 0.038 |
| MyVoteSys <$\alpha$=0.66, $\beta$=0.33, $\gamma$=0.80, MaxEntropy> | 2.858 | 0.436 | 0.038 |
| MyVoteSys <$\alpha$=0.66, $\beta$=____, $\gamma$=____, MinVariance> | 2.864 | 0.428 | 0.036 |
| MyVoteSys <$\alpha$=0.50, $\beta$=0.33, $\gamma$=0.80, Last> | 2.878 | 0.438 | 0.038 |
| MyVoteSys <$\alpha$=0.66, $\beta$=0.33, $\gamma$=0.80, Last> | 2.878 | 0.438 | 0.038 |
| MyVoteSys <$\alpha$=0.66, $\beta$=0.33, $\gamma$=0.80, MaxVariance> | 2.888 | 0.410 | 0.040 |

| | | | | | | |
|---|---|---|---|---|---|---|
| MyVoteSys <$\alpha$=0.66, $\beta$=____, $\gamma$=0.80, MaxVariance> | | | | 2.948 | 0.414 | 0.044 |
| MyVoteSys <$\alpha$=0.66, $\beta$=0.33, $\gamma$=____, MaxVariance> | | | | 2.976 | 0.412 | 0.040 |
| MyVoteSys <$\alpha$=0.66, $\beta$=____, $\gamma$=0.80, Last> | | | | 3.032 | 0.436 | 0.050 |
| MyVoteSys <$\alpha$=0.80, $\beta$=____, $\gamma$=____, First> | | | | 3.032 | 0.436 | 0.050 |
| MyVoteSys <$\alpha$=0.50, $\beta$=____, $\gamma$=0.80, Last> | | | | 3.032 | 0.436 | 0.050 |
| MyVoteSys <$\alpha$=0.50, $\beta$=____, $\gamma$=0.80, MaxEntropy> | | | | 3.032 | 0.434 | 0.050 |
| MyVoteSys <$\alpha$=0.66, $\beta$=____, $\gamma$=0.80, MaxEntropy> | | | | 3.032 | 0.434 | 0.050 |
| MyVoteSys <$\alpha$=0.80, $\beta$=____, $\gamma$=____, MinEntropy> | | | | 3.048 | 0.434 | 0.048 |
| MyVoteSys <$\alpha$=0.50, $\beta$=0.33, $\gamma$=____, MaxEntropy> | | | | 3.084 | 0.420 | 0.036 |
| MyVoteSys <$\alpha$=0.66, $\beta$=0.33, $\gamma$=____, MaxEntropy> | | | | 3.084 | 0.420 | 0.036 |
| MyVoteSys <$\alpha$=0.50, $\beta$=____, $\gamma$=0.66, MaxVariance> | | | | 3.096 | 0.352 | 0.022 |
| MyVoteSys <$\alpha$=0.50, $\beta$=0.33, $\gamma$=0.66, MaxVariance> | | | | 3.096 | 0.352 | 0.022 |
| MyVoteSys <$\alpha$=0.50, $\beta$=0.33, $\gamma$=____, Last> | | | | 3.102 | 0.420 | 0.036 |
| MyVoteSys <$\alpha$=0.66, $\beta$=0.33, $\gamma$=____, Last> | | | | 3.102 | 0.420 | 0.036 |
| MyVoteSys <$\alpha$=0.50, $\beta$=____, $\gamma$=0.66, MinEntropy> | | | | 3.120 | 0.362 | 0.020 |
| MyVoteSys <$\alpha$=0.50, $\beta$=____, $\gamma$=0.80, First> | | | | 3.120 | 0.362 | 0.020 |
| MyVoteSys <$\alpha$=0.50, $\beta$=0.33, $\gamma$=____, First> | | | | 3.120 | 0.362 | 0.020 |
| MyVoteSys <$\alpha$=0.50, $\beta$=____, $\gamma$=0.80, MinEntropy> | | | | 3.120 | 0.362 | 0.020 |
| MyVoteSys <$\alpha$=0.50, $\beta$=0.33, $\gamma$=0.80, First> | | | | 3.120 | 0.362 | 0.020 |
| MyVoteSys <$\alpha$=0.50, $\beta$=0.33, $\gamma$=____, MinEntropy> | | | | 3.120 | 0.362 | 0.020 |
| MyVoteSys <$\alpha$=0.50, $\beta$=____, $\gamma$=0.66, First> | | | | 3.120 | 0.362 | 0.020 |
| MyVoteSys <$\alpha$=0.50, $\beta$=____, $\gamma$=____, First> | | | | 3.120 | 0.362 | 0.020 |
| MyVoteSys <$\alpha$=0.50, $\beta$=0.33, $\gamma$=0.66, First> | | | | 3.120 | 0.362 | 0.020 |
| MyVoteSys <$\alpha$=0.50, $\beta$=0.33, $\gamma$=0.80, MinEntropy> | | | | 3.120 | 0.362 | 0.020 |
| MyVoteSys <$\alpha$=0.50, $\beta$=____, $\gamma$=____, MinEntropy> | | | | 3.120 | 0.362 | 0.020 |
| MyVoteSys <$\alpha$=0.50, $\beta$=0.33, $\gamma$=0.66, MinEntropy> | | | | 3.120 | 0.362 | 0.020 |
| MyVoteSys <$\alpha$=0.50, $\beta$=0.33, $\gamma$=____, MaxVariance> | | | | 3.170 | 0.350 | 0.026 |
| MyVoteSys <$\alpha$=0.50, $\beta$=0.33, $\gamma$=0.80, MaxVariance> | | | | 3.240 | 0.340 | 0.028 |
| MyVoteSys <$\alpha$=0.50, $\beta$=____, $\gamma$=0.80, MaxVariance> | | | | 3.290 | 0.340 | 0.030 |
| MyVoteSys <$\alpha$=0.80, $\beta$=____, $\gamma$=____, MinVariance> | | | | 3.446 | 0.410 | 0.068 |
| MyVoteSys <$\alpha$=0.80, $\beta$=0.33, $\gamma$=____, First> | | | | 3.812 | 0.396 | 0.022 |
| MyVoteSys <$\alpha$=0.80, $\beta$=0.33, $\gamma$=____, MinEntropy> | | | | 3.820 | 0.394 | 0.022 |
| MyVoteSys <$\alpha$=0.80, $\beta$=0.33, $\gamma$=____, MinVariance> | | | | 3.900 | 0.394 | 0.020 |
| MyVoteSys <$\alpha$=0.80, $\beta$=0.33, $\gamma$=____, MaxVariance> | | | | 3.960 | 0.384 | 0.022 |
| MyVoteSys <$\alpha$=0.80, $\beta$=0.33, $\gamma$=____, Last> | | | | 4.036 | 0.378 | 0.020 |
| MyVoteSys <$\alpha$=0.80, $\beta$=0.33, $\gamma$=____, MaxEntropy> | | | | 4.036 | 0.378 | 0.020 |
| PreferentialBlockVoting | | | | 6.244 | 0.204 | 0.218 |
| InstantRunoffVoting | | | | 6.254 | 0.202 | 0.218 |
| SingleTransferableVote | | | | 6.254 | 0.202 | 0.218 |
| MyVoteSys <$\alpha$=0.50, $\beta$=____, $\gamma$=____, MaxVariance> | | | | 8.026 | 0.096 | 0.322 |
| MyVoteSys <$\alpha$=0.66, $\beta$=____, $\gamma$=____, MaxVariance> | | | | 8.060 | 0.096 | 0.324 |
| MyVoteSys <$\alpha$=0.80, $\beta$=____, $\gamma$=____, MaxVariance> | | | | 8.116 | 0.094 | 0.328 |
| MyVoteSys <$\alpha$=0.66, $\beta$=____, $\gamma$=____, Last> | | | | 8.218 | 0.082 | 0.334 |
| MyVoteSys <$\alpha$=0.80, $\beta$=____, $\gamma$=____, MaxEntropy> | | | | 8.218 | 0.082 | 0.334 |
| MyVoteSys <$\alpha$=0.80, $\beta$=____, $\gamma$=____, Last> | | | | 8.218 | 0.082 | 0.334 |
| MyVoteSys <$\alpha$=0.50, $\beta$=____, $\gamma$=____, Last> | | | | 8.218 | 0.082 | 0.334 |
| MyVoteSys <$\alpha$=0.66, $\beta$=____, $\gamma$=____, MaxEntropy> | | | | 8.218 | 0.082 | 0.334 |
| MyVoteSys <$\alpha$=0.50, $\beta$=____, $\gamma$=____, MaxEntropy> | | | | 8.218 | 0.082 | 0.334 |
| FirstPastThePost (but without tactical voting) | | | | 8.236 | 0.120 | 0.354 |
| bestVoter (Dictatorship) | | | | 8.334 | 0.108 | 0.352 |

| Metrics | val_MeanSquaredErr |
|---|---|
| Algorithms | |
| crowd-Mean | 18604.785156 |
| crowd-Median | 18604.031250 |
| bestVoter (Dictatorship) | 16874.865234 |

============================= END OF SIMULATION 1 =============================

============================== START OF SIMULATION 2 ==============================

| Metrics | meanWinnerRank | rateTrueWinners | rateWinner<NULL |
|---|---|---|---|
| Algorithms | | | |

numCandiates : 20
numVoters : 750
numElections : 750
columnBlindness : 9
epochs : 133
trainableLayerCount : 1
crowdBuildMethod : {'name': 'standardDistribution', 'mean': 18750, 'standardDeviation': 266}
dataSetName : mySynthetic
predictedFeature : y

========= SIMULATION RESULTS ========

| Metrics | meanWinnerRank | rateTrueWinners | rateWinner<NULL |
|---|---|---|---|
| Algorithms | | | |
| crowd-Mean | 2.366 | 0.470 | 0.005 |
| crowd-Median | 2.366 | 0.470 | 0.005 |
| MyVoteSys <$\alpha$=0.50, $\beta$=____, $\gamma$=0.80, MinEntropy> | 6.717 | 0.165 | 0.234 |
| MyVoteSys <$\alpha$=0.50, $\beta$=____, $\gamma$=0.66, MinEntropy> | 6.717 | 0.165 | 0.234 |
| MyVoteSys <$\alpha$=0.50, $\beta$=0.33, $\gamma$=____, MinEntropy> | 6.717 | 0.165 | 0.234 |
| MyVoteSys <$\alpha$=0.50, $\beta$=0.33, $\gamma$=0.80, MinEntropy> | 6.717 | 0.165 | 0.234 |
| MyVoteSys <$\alpha$=0.50, $\beta$=0.33, $\gamma$=0.66, MinEntropy> | 6.717 | 0.165 | 0.234 |
| MyVoteSys <$\alpha$=0.50, $\beta$=____, $\gamma$=____, MinEntropy> | 6.717 | 0.165 | 0.234 |
| MyVoteSys <$\alpha$=0.50, $\beta$=____, $\gamma$=____, First > | 6.721 | 0.164 | 0.234 |
| MyVoteSys <$\alpha$=0.50, $\beta$=____, $\gamma$=0.80, First > | 6.721 | 0.164 | 0.234 |
| MyVoteSys <$\alpha$=0.50, $\beta$=____, $\gamma$=0.66, First > | 6.721 | 0.164 | 0.234 |
| MyVoteSys <$\alpha$=0.50, $\beta$=0.33, $\gamma$=____, First > | 6.721 | 0.164 | 0.234 |
| MyVoteSys <$\alpha$=0.50, $\beta$=0.33, $\gamma$=0.80, First > | 6.721 | 0.164 | 0.234 |
| MyVoteSys <$\alpha$=0.50, $\beta$=0.33, $\gamma$=0.66, First > | 6.721 | 0.164 | 0.234 |
| MyVoteSys <$\alpha$=0.50, $\beta$=____, $\gamma$=0.80, MaxStDev> | 7.701 | 0.125 | 0.300 |
| MyVoteSys <$\alpha$=0.50, $\beta$=0.33, $\gamma$=0.80, MaxStDev> | 7.701 | 0.125 | 0.300 |
| MyVoteSys <$\alpha$=0.50, $\beta$=____, $\gamma$=0.66, MaxVariance> | 7.701 | 0.125 | 0.300 |
| MyVoteSys <$\alpha$=0.50, $\beta$=0.33, $\gamma$=____, MaxStDev> | 7.701 | 0.125 | 0.300 |
| MyVoteSys <$\alpha$=0.50, $\beta$=____, $\gamma$=____, MaxStDev> | 7.701 | 0.125 | 0.300 |
| MyVoteSys <$\alpha$=0.50, $\beta$=____, $\gamma$=0.80, MaxVariance> | 7.701 | 0.125 | 0.300 |
| MyVoteSys <$\alpha$=0.50, $\beta$=0.33, $\gamma$=0.66, MaxVariance> | 7.701 | 0.125 | 0.300 |
| MyVoteSys <$\alpha$=0.50, $\beta$=____, $\gamma$=0.66, MaxStDev> | 7.701 | 0.125 | 0.300 |
| MyVoteSys <$\alpha$=0.50, $\beta$=____, $\gamma$=____, MaxVariance> | 7.701 | 0.125 | 0.300 |
| MyVoteSys <$\alpha$=0.50, $\beta$=0.33, $\gamma$=0.80, MaxVariance> | 7.701 | 0.125 | 0.300 |
| MyVoteSys <$\alpha$=0.50, $\beta$=0.33, $\gamma$=0.66, MaxStDev> | 7.701 | 0.125 | 0.300 |
| MyVoteSys <$\alpha$=0.50, $\beta$=0.33, $\gamma$=____, MaxVariance> | 7.701 | 0.125 | 0.300 |
| MaxStandardDeviationWeightedSumWins<$\alpha$=0> | 8.320 | 0.085 | 0.365 |
| MyVoteSys <$\alpha$=0.50, $\beta$=0.33, $\gamma$=0.66, MinVariance> | 8.456 | 0.108 | 0.366 |
| MyVoteSys <$\alpha$=0.50, $\beta$=____, $\gamma$=0.66, MinVariance> | 8.456 | 0.108 | 0.366 |
| bestVoter (Dictatorship) | 8.580 | 0.090 | 0.372 |
| MaxVarianceWeightedSumWins<$\alpha$=0> | 8.616 | 0.078 | 0.390 |
| MaxStandardDeviationWeightedSumWins<$\alpha$=0.5> | 8.941 | 0.064 | 0.413 |
| MaxVarianceWeightedSumWins<$\alpha$=0.5> | 8.974 | 0.068 | 0.418 |
| MaxStandardDeviationProportionalWeightedSumWins <$\alpha$=0> | 9.216 | 0.052 | 0.432 |
| MyVoteSys <$\alpha$=0.66, $\beta$=0.33, $\gamma$=0.80, First > | 9.230 | 0.082 | 0.429 |
| MyVoteSys <$\alpha$=0.50, $\beta$=____, $\gamma$=0.66, Last> | 9.230 | 0.082 | 0.429 |
| MyVoteSys <$\alpha$=0.66, $\beta$=0.33, $\gamma$=____, First > | 9.230 | 0.082 | 0.429 |
| MyVoteSys <$\alpha$=0.66, $\beta$=____, $\gamma$=0.80, First > | 9.230 | 0.082 | 0.429 |
| MyVoteSys <$\alpha$=0.50, $\beta$=0.33, $\gamma$=0.66, Last> | 9.230 | 0.082 | 0.429 |
| MyVoteSys <$\alpha$=0.66, $\beta$=____, $\gamma$=____, First > | 9.230 | 0.082 | 0.429 |
| MyVoteSys <$\alpha$=0.66, $\beta$=0.33, $\gamma$=0.80, MaxStDev> | 9.237 | 0.084 | 0.433 |
| MyVoteSys <$\alpha$=0.66, $\beta$=____, $\gamma$=0.80, MaxVariance> | 9.237 | 0.084 | 0.433 |
| MyVoteSys <$\alpha$=0.66, $\beta$=0.33, $\gamma$=0.80, MaxVariance> | 9.237 | 0.084 | 0.433 |
| MyVoteSys <$\alpha$=0.66, $\beta$=____, $\gamma$=0.80, MaxStDev> | 9.237 | 0.084 | 0.433 |
| MyVoteSys <$\alpha$=0.66, $\beta$=____, $\gamma$=____, MaxVariance> | 9.242 | 0.082 | 0.432 |
| MyVoteSys <$\alpha$=0.66, $\beta$=0.33, $\gamma$=____, MaxStDev> | 9.242 | 0.082 | 0.432 |
| MyVoteSys <$\alpha$=0.66, $\beta$=0.33, $\gamma$=____, MaxVariance> | 9.242 | 0.082 | 0.432 |
| MyVoteSys <$\alpha$=0.66, $\beta$=____, $\gamma$=____, MaxStDev> | 9.242 | 0.082 | 0.432 |

| | val_MeanSquaredErr | | |
|---|---|---|---|
| MyVoteSys <$\alpha$=0.66, $\beta$=____, $\gamma$=0.80, MinEntropy> | 9.258 | 0.081 | 0.434 |
| MyVoteSys <$\alpha$=0.66, $\beta$=0.33, $\gamma$=0.80, MinEntropy> | 9.258 | 0.081 | 0.434 |
| MaxVarianceProportionalWeightedSumWins<$\alpha$=0> | 9.258 | 0.052 | 0.438 |
| MyVoteSys <$\alpha$=0.66, $\beta$=0.33, $\gamma$=____, MinEntropy> | 9.264 | 0.080 | 0.434 |
| MyVoteSys <$\alpha$=0.66, $\beta$=____, $\gamma$=____, MinEntropy> | 9.264 | 0.080 | 0.434 |
| MyVoteSys <$\alpha$=0.50, $\beta$=____, $\gamma$=0.66, MaxEntropy> | 9.269 | 0.081 | 0.433 |
| MyVoteSys <$\alpha$=0.50, $\beta$=0.33, $\gamma$=0.66, MaxEntropy> | 9.269 | 0.081 | 0.433 |
| MinEntropyWeightedSumWins2<$\alpha$=0> | 9.328 | 0.088 | 0.428 |
| MinEntropyProportionalWeightedSumWins2<$\alpha$=0> | 9.328 | 0.088 | 0.428 |
| MinEntropyWeightedSumWins<$\alpha$=0> | 9.328 | 0.088 | 0.428 |
| MinEntropyProportionalWeightedSumWins <$\alpha$=0> | 9.328 | 0.088 | 0.428 |
| MaxStandardDeviationProportionalWeightedSumWin<$\alpha$=0.5>| 9.336 | 0.048 | 0.442 |
| MinEntropyWeightedSumWins2<$\alpha$=0.5> | 9.349 | 0.088 | 0.429 |
| MinEntropyProportionalWeightedSumWins <$\alpha$=0.5> | 9.349 | 0.088 | 0.429 |
| MinEntropyProportionalWeightedSumWins2<$\alpha$=0.5> | 9.349 | 0.088 | 0.429 |
| MinEntropyWeightedSumWins<$\alpha$=0.5> | 9.349 | 0.088 | 0.429 |
| MaxVarianceProportionalWeightedSumWins<$\alpha$=0.5> | 9.362 | 0.050 | 0.448 |
| MyVoteSys <$\alpha$=0.50, $\beta$=0.33, $\gamma$=0.80, MinVariance> | 9.393 | 0.058 | 0.461 |
| MyVoteSys <$\alpha$=0.50, $\beta$=____, $\gamma$=0.80, MinVariance> | 9.393 | 0.058 | 0.461 |
| MyVoteSys <$\alpha$=0.50, $\beta$=____, $\gamma$=0.80, MaxEntropy> | 9.409 | 0.056 | 0.460 |
| MyVoteSys <$\alpha$=0.50, $\beta$=0.33, $\gamma$=0.80, MaxEntropy> | 9.409 | 0.056 | 0.460 |
| MyVoteSys <$\alpha$=0.66, $\beta$=____, $\gamma$=0.80, MaxEntropy> | 9.412 | 0.056 | 0.460 |
| MyVoteSys <$\alpha$=0.66, $\beta$=0.33, $\gamma$=0.80, MaxEntropy> | 9.412 | 0.056 | 0.460 |
| MyVoteSys <$\alpha$=0.50, $\beta$=____, $\gamma$=0.80, Last> | 9.448 | 0.056 | 0.465 |
| MyVoteSys <$\alpha$=0.50, $\beta$=0.33, $\gamma$=0.80, Last> | 9.448 | 0.056 | 0.465 |
| MyVoteSys <$\alpha$=0.80, $\beta$=____, $\gamma$=____, First > | 9.448 | 0.056 | 0.465 |
| MyVoteSys <$\alpha$=0.66, $\beta$=0.33, $\gamma$=0.80, Last> | 9.448 | 0.056 | 0.465 |
| MyVoteSys <$\alpha$=0.66, $\beta$=____, $\gamma$=0.80, Last> | 9.448 | 0.056 | 0.465 |
| MyVoteSys <$\alpha$=0.80, $\beta$=0.33, $\gamma$=____, First > | 9.448 | 0.056 | 0.465 |
| MyVoteSys <$\alpha$=0.80, $\beta$=____, $\gamma$=____, MinEntropy> | 9.450 | 0.056 | 0.464 |
| MyVoteSys <$\alpha$=0.80, $\beta$=0.33, $\gamma$=____, MinEntropy> | 9.450 | 0.056 | 0.464 |
| MyVoteSys <$\alpha$=0.66, $\beta$=____, $\gamma$=0.80, MinVariance> | 9.485 | 0.056 | 0.466 |
| MyVoteSys <$\alpha$=0.66, $\beta$=0.33, $\gamma$=0.80, MinVariance> | 9.485 | 0.056 | 0.466 |
| MyVoteSys <$\alpha$=0.50, $\beta$=____, $\gamma$=____, MinVariance> | 9.581 | 0.056 | 0.473 |
| MyVoteSys <$\alpha$=0.50, $\beta$=0.33, $\gamma$=____, MinVariance> | 9.581 | 0.056 | 0.473 |
| MyVoteSys <$\alpha$=0.80, $\beta$=____, $\gamma$=____, MinVariance> | 9.605 | 0.053 | 0.474 |
| MyVoteSys <$\alpha$=0.66, $\beta$=____, $\gamma$=____, MinVariance> | 9.605 | 0.053 | 0.474 |
| MyVoteSys <$\alpha$=0.66, $\beta$=0.33, $\gamma$=____, MinVariance> | 9.605 | 0.053 | 0.474 |
| MyVoteSys <$\alpha$=0.80, $\beta$=0.33, $\gamma$=____, MinVariance> | 9.605 | 0.053 | 0.474 |
| MyVoteSys <$\alpha$=0.80, $\beta$=____, $\gamma$=____, MaxVariance> | 9.821 | 0.033 | 0.470 |
| MyVoteSys <$\alpha$=0.80, $\beta$=0.33, $\gamma$=____, MaxVariance> | 9.821 | 0.033 | 0.470 |
| MyVoteSys <$\alpha$=0.80, $\beta$=0.33, $\gamma$=____, MaxStDev> | 9.821 | 0.033 | 0.470 |
| MyVoteSys <$\alpha$=0.80, $\beta$=____, $\gamma$=____, MaxStDev> | 9.821 | 0.033 | 0.470 |
| MyVoteSys <$\alpha$=0.50, $\beta$=____, $\gamma$=____, Last> | 9.886 | 0.030 | 0.474 |
| MyVoteSys <$\alpha$=0.50, $\beta$=____, $\gamma$=____, MaxEntropy> | 9.886 | 0.030 | 0.474 |
| MyVoteSys <$\alpha$=0.80, $\beta$=____, $\gamma$=____, MaxEntropy> | 9.886 | 0.030 | 0.474 |
| MyVoteSys <$\alpha$=0.80, $\beta$=0.33, $\gamma$=____, MaxEntropy> | 9.886 | 0.030 | 0.474 |
| MyVoteSys <$\alpha$=0.66, $\beta$=____, $\gamma$=____, MaxEntropy> | 9.886 | 0.030 | 0.474 |
| MyVoteSys <$\alpha$=0.66, $\beta$=0.33, $\gamma$=____, MaxEntropy> | 9.886 | 0.030 | 0.474 |
| MyVoteSys <$\alpha$=0.80, $\beta$=____, $\gamma$=____, Last> | 9.886 | 0.030 | 0.474 |
| MyVoteSys <$\alpha$=0.66, $\beta$=____, $\gamma$=____, Last> | 9.886 | 0.030 | 0.474 |
| MyVoteSys <$\alpha$=0.80, $\beta$=0.33, $\gamma$=____, Last> | 9.886 | 0.030 | 0.474 |
| MyVoteSys <$\alpha$=0.66, $\beta$=0.33, $\gamma$=____, Last> | 9.886 | 0.030 | 0.474 |
| MyVoteSys <$\alpha$=0.50, $\beta$=0.33, $\gamma$=____, Last> | 9.886 | 0.030 | 0.474 |
| MyVoteSys <$\alpha$=0.50, $\beta$=0.33, $\gamma$=____, MaxEntropy> | 9.886 | 0.030 | 0.474 |
| SingleTransferableVote | 0.630 | 0.049 | 0.521 |
| PreferentialBlockVoting | 0.632 | 0.049 | 0.521 |
| InstantRunoffVoting | 0.632 | 0.049 | 0.521 |
| FirstPastThePost (but without tactical voting) | 0.742 | 0.048 | 0.528 |

Metrics                        val_MeanSquaredErr

Algorithms
crowd–Mean                         18409.195312
crowd–Median                       18410.566406
bestVoter (Dictatorship)           17580.296875
================================ END OF SIMULATION 2 ================================

================================ START OF SIMULATION 3 ================================
numCandiates : 20
numVoters : 750
numElections : 750
columnBlindness : 8
epochs : 166
trainableLayerCount : 1
crowdBuildMethod : {'name': 'standardDistribution', 'mean': 16000, 'standardDeviation': 666}
dataSetName : mySynthetic
predictedFeature : y
========= SIMULATION RESULTS ========

| Metrics | meanWinnerRank | rateTrueWinners | rateWinner<NULL |
|---|---|---|---|
| Algorithms | | | |
| crowd–Median | 1.144 | 0.877 | 0.0000 |
| crowd–Mean | 1.144 | 0.877 | 0.0000 |
| MyVoteSys <$\alpha$=0.50, $\beta$=0.33, $\gamma$=0.66, MaxEntropy> | 1.360 | 0.744 | 0.0000 |
| MyVoteSys <$\alpha$=0.50, $\beta$=0.33, $\gamma$=0.66, Last> | 1.360 | 0.744 | 0.0000 |
| MyVoteSys <$\alpha$=0.50, $\beta$=0.33, $\gamma$=0.66, MinVariance> | 1.369 | 0.737 | 0.0000 |
| MyVoteSys <$\alpha$=0.66, $\beta$=____, $\gamma$=____, MinEntropy> | 1.370 | 0.741 | 0.0000 |
| MyVoteSys <$\alpha$=0.50, $\beta$=____, $\gamma$=0.66, MaxEntropy> | 1.370 | 0.741 | 0.0000 |
| MyVoteSys <$\alpha$=0.66, $\beta$=____, $\gamma$=0.80, MinEntropy> | 1.370 | 0.741 | 0.0000 |
| MyVoteSys <$\alpha$=0.66, $\beta$=____, $\gamma$=0.80, First > | 1.370 | 0.741 | 0.0000 |
| MyVoteSys <$\alpha$=0.50, $\beta$=____, $\gamma$=0.66, Last> | 1.370 | 0.741 | 0.0000 |
| MyVoteSys <$\alpha$=0.66, $\beta$=____, $\gamma$=____, First > | 1.370 | 0.741 | 0.0000 |
| MyVoteSys <$\alpha$=0.50, $\beta$=____, $\gamma$=0.66, MinVariance> | 1.372 | 0.737 | 0.0000 |
| MyVoteSys <$\alpha$=0.50, $\beta$=0.33, $\gamma$=0.80, MinEntropy> | 1.412 | 0.712 | 0.0000 |
| MyVoteSys <$\alpha$=0.50, $\beta$=____, $\gamma$=0.80, First > | 1.412 | 0.712 | 0.0000 |
| MyVoteSys <$\alpha$=0.50, $\beta$=____, $\gamma$=0.80, MinEntropy> | 1.412 | 0.712 | 0.0000 |
| MyVoteSys <$\alpha$=0.50, $\beta$=0.33, $\gamma$=0.80, First > | 1.412 | 0.712 | 0.0000 |
| MyVoteSys <$\alpha$=0.50, $\beta$=0.33, $\gamma$=0.66, First > | 1.412 | 0.712 | 0.0000 |
| MyVoteSys <$\alpha$=0.50, $\beta$=____, $\gamma$=0.66, First > | 1.412 | 0.712 | 0.0000 |
| MyVoteSys <$\alpha$=0.50, $\beta$=0.33, $\gamma$=____, MinEntropy> | 1.412 | 0.712 | 0.0000 |
| MyVoteSys <$\alpha$=0.50, $\beta$=____, $\gamma$=0.66, MinEntropy> | 1.412 | 0.712 | 0.0000 |
| MyVoteSys <$\alpha$=0.50, $\beta$=0.33, $\gamma$=0.66, MinEntropy> | 1.412 | 0.712 | 0.0000 |
| MyVoteSys <$\alpha$=0.50, $\beta$=0.33, $\gamma$=____, First > | 1.412 | 0.712 | 0.0000 |
| MyVoteSys <$\alpha$=0.50, $\beta$=____, $\gamma$=____, MinEntropy> | 1.412 | 0.712 | 0.0000 |
| MyVoteSys <$\alpha$=0.50, $\beta$=0.33, $\gamma$=0.80, MaxVariance> | 1.413 | 0.710 | 0.0000 |
| MyVoteSys <$\alpha$=0.50, $\beta$=____, $\gamma$=0.66, MaxVariance> | 1.413 | 0.710 | 0.0000 |
| MyVoteSys <$\alpha$=0.50, $\beta$=0.33, $\gamma$=____, MaxVariance> | 1.413 | 0.710 | 0.0000 |
| MyVoteSys <$\alpha$=0.50, $\beta$=0.33, $\gamma$=0.66, MaxVariance> | 1.413 | 0.710 | 0.0000 |
| MyVoteSys <$\alpha$=0.50, $\beta$=0.33, $\gamma$=0.80, Last> | 1.424 | 0.724 | 0.0000 |
| MyVoteSys <$\alpha$=0.50, $\beta$=0.33, $\gamma$=0.80, MaxEntropy> | 1.424 | 0.724 | 0.0000 |
| MyVoteSys <$\alpha$=0.50, $\beta$=____, $\gamma$=0.80, MaxVariance> | 1.424 | 0.709 | 0.0000 |
| MyVoteSys <$\alpha$=0.50, $\beta$=0.33, $\gamma$=0.80, MinVariance> | 1.428 | 0.713 | 0.0000 |
| MyVoteSys <$\alpha$=0.66, $\beta$=____, $\gamma$=0.80, MinVariance> | 1.436 | 0.714 | 0.0000 |
| MyVoteSys <$\alpha$=0.50, $\beta$=____, $\gamma$=0.80, MinVariance> | 1.437 | 0.710 | 0.0000 |
| MyVoteSys <$\alpha$=0.50, $\beta$=0.33, $\gamma$=____, MinVariance> | 1.437 | 0.705 | 0.0000 |
| MyVoteSys <$\alpha$=0.50, $\beta$=0.33, $\gamma$=____, Last> | 1.441 | 0.714 | 0.0000 |
| MyVoteSys <$\alpha$=0.50, $\beta$=0.33, $\gamma$=____, MaxEntropy> | 1.441 | 0.714 | 0.0000 |
| MyVoteSys <$\alpha$=0.66, $\beta$=0.33, $\gamma$=____, MinEntropy> | 1.444 | 0.737 | 0.0000 |
| MyVoteSys <$\alpha$=0.66, $\beta$=0.33, $\gamma$=0.80, First > | 1.444 | 0.737 | 0.0000 |
| MyVoteSys <$\alpha$=0.66, $\beta$=0.33, $\gamma$=0.80, MinEntropy> | 1.444 | 0.737 | 0.0000 |
| MyVoteSys <$\alpha$=0.66, $\beta$=0.33, $\gamma$=____, First > | 1.444 | 0.737 | 0.0000 |
| MyVoteSys <$\alpha$=0.66, $\beta$=____, $\gamma$=____, MinVariance> | 1.445 | 0.706 | 0.0000 |
| MyVoteSys <$\alpha$=0.50, $\beta$=____, $\gamma$=____, MinVariance> | 1.446 | 0.702 | 0.0000 |
| MyVoteSys <$\alpha$=0.66, $\beta$=____, $\gamma$=0.80, MaxVariance> | 1.449 | 0.725 | 0.0013 |
| MyVoteSys <$\alpha$=0.66, $\beta$=0.33, $\gamma$=0.80, MaxVariance> | 1.450 | 0.738 | 0.0013 |
| MyVoteSys <$\alpha$=0.66, $\beta$=0.33, $\gamma$=____, MaxVariance> | 1.452 | 0.738 | 0.0000 |
| MyVoteSys <$\alpha$=0.50, $\beta$=____, $\gamma$=0.80, Last> | 1.501 | 0.702 | 0.0013 |
| MyVoteSys <$\alpha$=0.66, $\beta$=____, $\gamma$=0.80, Last> | 1.501 | 0.702 | 0.0013 |
| MyVoteSys <$\alpha$=0.50, $\beta$=____, $\gamma$=0.80, MaxEntropy> | 1.501 | 0.702 | 0.0013 |
| MyVoteSys <$\alpha$=0.66, $\beta$=____, $\gamma$=0.80, MaxEntropy> | 1.501 | 0.702 | 0.0013 |
| MyVoteSys <$\alpha$=0.80, $\beta$=____, $\gamma$=____, MinEntropy> | 1.501 | 0.702 | 0.0013 |
| MyVoteSys <$\alpha$=0.80, $\beta$=____, $\gamma$=____, First> | 1.501 | 0.702 | 0.0013 |
| MyVoteSys <$\alpha$=0.66, $\beta$=0.33, $\gamma$=0.80, MinVariance> | 1.502 | 0.713 | 0.0000 |
| MyVoteSys <$\alpha$=0.66, $\beta$=0.33, $\gamma$=0.80, Last> | 1.508 | 0.717 | 0.0000 |
| MyVoteSys <$\alpha$=0.66, $\beta$=0.33, $\gamma$=0.80, MaxEntropy> | 1.508 | 0.717 | 0.0000 |
| MyVoteSys <$\alpha$=0.66, $\beta$=0.33, $\gamma$=____, MinVariance> | 1.512 | 0.705 | 0.0000 |
| MyVoteSys <$\alpha$=0.80, $\beta$=____, $\gamma$=____, MinVariance> | 1.512 | 0.694 | 0.0013 |
| MyVoteSys <$\alpha$=0.66, $\beta$=0.33, $\gamma$=____, MaxEntropy> | 1.525 | 0.708 | 0.0000 |
| MyVoteSys <$\alpha$=0.66, $\beta$=0.33, $\gamma$=____, Last> | 1.525 | 0.708 | 0.0000 |
| PreferentialBlockVoting | 1.574 | 0.665 | 0.0000 |
| SingleTransferableVote | 1.580 | 0.664 | 0.0000 |
| InstantRunoffVoting | 1.581 | 0.664 | 0.0000 |
| FirstPastThePost (but without tactical voting) | 2.605 | 0.470 | 0.0240 |
| MyVoteSys <$\alpha$=0.50, $\beta$=____, $\gamma$=____, MaxVariance> | 3.180 | 0.402 | 0.0480 |
| MyVoteSys <$\alpha$=0.66, $\beta$=____, $\gamma$=____, MaxVariance> | 3.180 | 0.402 | 0.0480 |
| MyVoteSys <$\alpha$=0.80, $\beta$=____, $\gamma$=____, MaxVariance> | 3.180 | 0.402 | 0.0480 |
| MyVoteSys <$\alpha$=0.80, $\beta$=____, $\gamma$=____, MaxEntropy> | 3.180 | 0.402 | 0.0480 |
| MyVoteSys <$\alpha$=0.50, $\beta$=____, $\gamma$=____, MaxEntropy> | 3.180 | 0.402 | 0.0480 |
| MyVoteSys <$\alpha$=0.50, $\beta$=____, $\gamma$=____, Last> | 3.180 | 0.402 | 0.0480 |
| MyVoteSys <$\alpha$=0.80, $\beta$=____, $\gamma$=____, Last> | 3.180 | 0.402 | 0.0480 |
| MyVoteSys <$\alpha$=0.66, $\beta$=____, $\gamma$=____, MaxEntropy> | 3.180 | 0.402 | 0.0480 |
| MyVoteSys <$\alpha$=0.66, $\beta$=____, $\gamma$=____, Last> | 3.180 | 0.402 | 0.0480 |
| MyVoteSys <$\alpha$=0.80, $\beta$=0.33, $\gamma$=____, MinEntropy> | 4.164 | 0.530 | 0.0000 |
| MyVoteSys <$\alpha$=0.80, $\beta$=0.33, $\gamma$=____, First> | 4.164 | 0.530 | 0.0000 |
| MyVoteSys <$\alpha$=0.80, $\beta$=0.33, $\gamma$=____, MinVariance> | 4.173 | 0.522 | 0.0000 |
| MyVoteSys <$\alpha$=0.80, $\beta$=0.33, $\gamma$=____, MaxVariance> | 4.180 | 0.522 | 0.0000 |
| MyVoteSys <$\alpha$=0.80, $\beta$=0.33, $\gamma$=____, MaxEntropy> | 4.181 | 0.521 | 0.0000 |
| MyVoteSys <$\alpha$=0.80, $\beta$=0.33, $\gamma$=____, Last> | 4.181 | 0.521 | 0.0000 |
| bestVoter (Dictatorship) | 5.284 | 0.208 | 0.1266 |

| Metrics | val_MeanSquaredErr |
|---|---|
| Algorithms | |
| crowd-Median | 17898.462891 |
| crowd-Mean | 17960.232422 |
| bestVoter (Dictatorship) | 14130.808594 |

================================= END OF SIMULATION 3 =================================

================================= START OF SIMULATION 4 =================================
numCandiates : 20
numVoters : 500
numElections : 1000
columnBlindness : 5
epochs : 66
trainableLayerCount : 1
crowdBuildMethod : {'name': 'standardDistribution', 'mean': 9500, 'standardDeviation': 1000}
dataSetName : mySynthetic
predictedFeature : y

========= SIMULATION RESULTS =========

| Metrics | meanWinnerRank | rateTrueWinners | rateWinner<NULL |
|---|---|---|---|
| Algorithms | | | |
| MyVoteSys <$\alpha$=0.50, $\beta$=____, $\gamma$=0.80, MinEntropy> | 1.219 | 0.837 | 0.000 |
| MyVoteSys <$\alpha$=0.50, $\beta$=0.33, $\gamma$=0.66, MaxVariance> | 1.219 | 0.837 | 0.000 |
| MyVoteSys <α=0.50, β=0.33, γ=0.80, MaxVariance> | 1.219 | 0.837 | 0.000 |
| MyVoteSys <α=0.50, β=____, γ=____, First> | 1.219 | 0.837 | 0.000 |
| MyVoteSys <α=0.50, β=____, γ=0.66, MinEntropy> | 1.219 | 0.837 | 0.000 |
| MyVoteSys <α=0.50, β=0.33, γ=____, MaxVariance> | 1.219 | 0.837 | 0.000 |
| MyVoteSys <α=0.50, β=____, γ=0.80, First> | 1.219 | 0.837 | 0.000 |
| MyVoteSys <α=0.50, β=0.33, γ=0.80, MinEntropy> | 1.219 | 0.837 | 0.000 |
| MyVoteSys <α=0.50, β=0.33, γ=____, MinEntropy> | 1.219 | 0.837 | 0.000 |
| MyVoteSys <α=0.50, β=____, γ=____, MinEntropy> | 1.219 | 0.837 | 0.000 |
| MyVoteSys <α=0.50, β=0.33, γ=____, First> | 1.219 | 0.837 | 0.000 |
| MyVoteSys <α=0.50, β=____, γ=0.66, First> | 1.219 | 0.837 | 0.000 |
| MyVoteSys <α=0.50, β=____, γ=0.66, MaxVariance> | 1.219 | 0.837 | 0.000 |
| MyVoteSys <α=0.50, β=____, γ=0.80, MaxVariance> | 1.219 | 0.837 | 0.000 |
| MyVoteSys <α=0.50, β=0.33, γ=0.66, MinEntropy> | 1.219 | 0.837 | 0.000 |
| MyVoteSys <α=0.50, β=0.33, γ=0.80, First> | 1.219 | 0.837 | 0.000 |
| MyVoteSys <α=0.50, β=0.33, γ=0.66, First> | 1.219 | 0.837 | 0.000 |
| crowd-Median | 1.220 | 0.839 | 0.000 |
| crowd-Mean | 1.220 | 0.839 | 0.000 |
| MyVoteSys <α=0.66, β=____, γ=0.80, MaxVariance> | 1.232 | 0.824 | 0.000 |
| MyVoteSys <α=0.66, β=____, γ=____, First> | 1.232 | 0.824 | 0.000 |
| MyVoteSys <α=0.66, β=____, γ=____, MinEntropy> | 1.232 | 0.824 | 0.000 |
| MyVoteSys <α=0.50, β=0.33, γ=0.66, MaxEntropy> | 1.232 | 0.824 | 0.000 |
| MyVoteSys <α=0.50, β=0.33, γ=0.66, MinVariance> | 1.232 | 0.824 | 0.000 |
| MyVoteSys <α=0.66, β=____, γ=0.80, MinEntropy> | 1.232 | 0.824 | 0.000 |
| MyVoteSys <α=0.50, β=____, γ=0.66, MaxEntropy> | 1.232 | 0.824 | 0.000 |
| MyVoteSys <α=0.50, β=____, γ=0.66, MinVariance> | 1.232 | 0.824 | 0.000 |
| MyVoteSys <α=0.66, β=0.33, γ=0.80, MinEntropy> | 1.232 | 0.824 | 0.000 |
| MyVoteSys <α=0.66, β=0.33, γ=0.80, First> | 1.232 | 0.824 | 0.000 |
| MyVoteSys <α=0.66, β=0.33, γ=____, MaxVariance> | 1.232 | 0.824 | 0.000 |
| MyVoteSys <α=0.50, β=____, γ=0.66, Last> | 1.232 | 0.824 | 0.000 |
| MyVoteSys <α=0.66, β=0.33, γ=0.80, MaxVariance> | 1.232 | 0.824 | 0.000 |
| MyVoteSys <α=0.66, β=____, γ=0.80, First> | 1.232 | 0.824 | 0.000 |
| MyVoteSys <α=0.66, β=0.33, γ=____, First> | 1.232 | 0.824 | 0.000 |
| MyVoteSys <α=0.50, β=0.33, γ=0.66, Last> | 1.232 | 0.824 | 0.000 |
| MyVoteSys <α=0.66, β=0.33, γ=____, MinEntropy> | 1.232 | 0.824 | 0.000 |
| PreferentialBlockVoting | 1.240 | 0.824 | 0.000 |
| SingleTransferableVote | 1.241 | 0.826 | 0.000 |
| InstantRunoffVoting | 1.241 | 0.823 | 0.000 |
| MyVoteSys <α=0.66, β=0.33, γ=0.80, MinVariance> | 1.297 | 0.793 | 0.000 |
| MyVoteSys <α=0.50, β=0.33, γ=0.80, MinVariance> | 1.297 | 0.793 | 0.000 |
| MyVoteSys <α=0.50, β=0.33, γ=0.80, Last> | 1.298 | 0.792 | 0.000 |
| MyVoteSys <α=0.66, β=0.33, γ=0.80, Last> | 1.298 | 0.792 | 0.000 |
| MyVoteSys <α=0.50, β=0.33, γ=0.80, MaxEntropy> | 1.298 | 0.792 | 0.000 |
| MyVoteSys <α=0.66, β=0.33, γ=0.80, MaxEntropy> | 1.298 | 0.792 | 0.000 |
| MyVoteSys <α=0.66, β=____, γ=0.80, MinVariance> | 1.304 | 0.790 | 0.000 |
| MyVoteSys <α=0.50, β=____, γ=0.80, MinVariance> | 1.304 | 0.790 | 0.000 |
| MyVoteSys <α=0.66, β=____, γ=0.80, MaxEntropy> | 1.305 | 0.789 | 0.000 |
| MyVoteSys <α=0.50, β=____, γ=0.80, MaxEntropy> | 1.305 | 0.789 | 0.000 |
| MyVoteSys <α=0.66, β=____, γ=0.80, Last> | 1.305 | 0.789 | 0.000 |
| MyVoteSys <α=0.80, β=____, γ=____, MinEntropy> | 1.305 | 0.789 | 0.000 |
| MyVoteSys <α=0.80, β=____, γ=____, First> | 1.305 | 0.789 | 0.000 |
| MyVoteSys <α=0.50, β=____, γ=0.80, Last> | 1.305 | 0.789 | 0.000 |
| MyVoteSys <α=0.50, β=0.33, γ=____, MinVariance> | 1.364 | 0.764 | 0.000 |
| MyVoteSys <α=0.66, β=0.33, γ=____, MinVariance> | 1.364 | 0.764 | 0.000 |
| MyVoteSys <α=0.50, β=0.33, γ=____, Last> | 1.390 | 0.752 | 0.000 |
| MyVoteSys <α=0.66, β=0.33, γ=____, Last> | 1.390 | 0.752 | 0.000 |
| MyVoteSys <α=0.66, β=0.33, γ=____, MaxEntropy> | 1.390 | 0.752 | 0.000 |
| MyVoteSys <α=0.50, β=0.33, γ=____, MaxEntropy> | 1.390 | 0.752 | 0.000 |
| FirstPastThePost (but without tactical voting) | 1.391 | 0.761 | 0.000 |
| MyVoteSys <α=0.80, β=____, γ=____, First> | 1.393 | 0.784 | 0.000 |
| MyVoteSys <α=0.80, β=0.33, γ=____, MinEntropy> | 1.393 | 0.784 | 0.000 |
| MyVoteSys <α=0.80, β=0.33, γ=____, MaxVariance> | 1.397 | 0.784 | 0.000 |
| MyVoteSys <$\alpha$=0.80, $\beta$=0.33, $\gamma$=____, MinVariance> | 1.460 | 0.755 | 0.000 |
| MyVoteSys <$\alpha$=0.80, $\beta$=0.33, $\gamma$=____, Last> | 1.485 | 0.744 | 0.000 |
| MyVoteSys <$\alpha$=0.80, $\beta$=0.33, $\gamma$=____, MaxEntropy> | 1.485 | 0.744 | 0.000 |
| MyVoteSys <$\alpha$=0.66, $\beta$=____, $\gamma$=____, MinVariance> | 1.647 | 0.721 | 0.000 |
| MyVoteSys <$\alpha$=0.80, $\beta$=____, $\gamma$=____, MinVariance> | 1.647 | 0.721 | 0.000 |
| MyVoteSys <$\alpha$=0.50, $\beta$=____, $\gamma$=____, MinVariance> | 1.647 | 0.721 | 0.000 |
| MyVoteSys <$\alpha$=0.50, $\beta$=____, $\gamma$=____, MaxVariance> | 2.609 | 0.623 | 0.000 |
| MyVoteSys <$\alpha$=0.66, $\beta$=____, $\gamma$=____, MaxVariance> | 2.690 | 0.605 | 0.000 |
| MyVoteSys <$\alpha$=0.80, $\beta$=____, $\gamma$=____, MaxVariance> | 2.690 | 0.605 | 0.000 |
| MyVoteSys <$\alpha$=0.66, $\beta$=____, $\gamma$=____, Last> | 2.698 | 0.604 | 0.000 |
| MyVoteSys <$\alpha$=0.80, $\beta$=____, $\gamma$=____, MaxEntropy> | 2.698 | 0.604 | 0.000 |
| MyVoteSys <$\alpha$=0.50, $\beta$=____, $\gamma$=____, Last> | 2.698 | 0.604 | 0.000 |
| MyVoteSys <$\alpha$=0.80, $\beta$=____, $\gamma$=____, Last> | 2.698 | 0.604 | 0.000 |
| MyVoteSys <$\alpha$=0.66, $\beta$=____, $\gamma$=____, MaxEntropy> | 2.698 | 0.604 | 0.000 |
| MyVoteSys <$\alpha$=0.50, $\beta$=____, $\gamma$=____, MaxEntropy> | 2.698 | 0.604 | 0.000 |
| bestVoter (Dictatorship) | 2.758 | 0.427 | 0.023 |

| Metrics | val_MeanSquaredErr |
|---|---|
| Algorithms | |
| crowd-Median | 21553.529297 |
| crowd-Mean | 21676.593750 |
| bestVoter (Dictatorship) | 6908.480957 |

=============================== END OF SIMULATION 4 ===============================

=============================== START OF SIMULATION 5 ===============================
numCandiates : 20
numVoters : 750
numElections : 750
columnBlindness : 5
epochs : 166
trainableLayerCount : 1
crowdBuildMethod : {'name': 'standardDistribution', 'mean': 9500, 'standardDeviation': 1000}
dataSetName : mySynthetic
predictedFeature : y

========= SIMULATION RESULTS ========

| Metrics | meanWinnerRank | rateTrueWinners | rateWinner<NULL |
|---|---|---|---|
| Algorithms | | | |
| crowd-Mean | 1.189 | 0.857 | 0.000 |
| crowd-Median | 1.189 | 0.857 | 0.000 |
| InstantRunoffVoting | 1.221 | 0.834 | 0.000 |
| MyVoteSys <$\alpha$=0.50, $\beta$=____, $\gamma$=____, MinEntropy> | 1.225 | 0.833 | 0.000 |
| MyVoteSys <$\alpha$=0.50, $\beta$=0.33, $\gamma$=____, MinEntropy> | 1.225 | 0.833 | 0.000 |
| MyVoteSys <$\alpha$=0.50, $\beta$=0.33, $\gamma$=0.80, MinEntropy> | 1.225 | 0.833 | 0.000 |
| MyVoteSys <$\alpha$=0.50, $\beta$=0.33, $\gamma$=0.66, MaxStDev> | 1.225 | 0.833 | 0.000 |
| MyVoteSys <$\alpha$=0.50, $\beta$=____, $\gamma$=0.80, First> | 1.225 | 0.833 | 0.000 |
| MyVoteSys <$\alpha$=0.50, $\beta$=____, $\gamma$=0.66, First> | 1.225 | 0.833 | 0.000 |
| MyVoteSys <$\alpha$=0.50, $\beta$=0.33, $\gamma$=0.66, MinEntropy> | 1.225 | 0.833 | 0.000 |
| MyVoteSys <$\alpha$=0.50, $\beta$=0.33, $\gamma$=0.66, First> | 1.225 | 0.833 | 0.000 |
| MyVoteSys <$\alpha$=0.50, $\beta$=____, $\gamma$=____, First> | 1.225 | 0.833 | 0.000 |
| MyVoteSys <$\alpha$=0.50, $\beta$=0.33, $\gamma$=0.80, MaxVariance> | 1.225 | 0.833 | 0.000 |
| MyVoteSys <$\alpha$=0.50, $\beta$=____, $\gamma$=0.66, MaxVariance> | 1.225 | 0.833 | 0.000 |
| MyVoteSys <$\alpha$=0.50, $\beta$=____, $\gamma$=0.66, MaxStDev> | 1.225 | 0.833 | 0.000 |
| MyVoteSys <$\alpha$=0.50, $\beta$=____, $\gamma$=0.80, MaxVariance> | 1.225 | 0.833 | 0.000 |
| MyVoteSys <$\alpha$=0.50, $\beta$=____, $\gamma$=0.66, MinEntropy> | 1.225 | 0.833 | 0.000 |
| MyVoteSys <$\alpha$=0.50, $\beta$=0.33, $\gamma$=0.80, First> | 1.225 | 0.833 | 0.000 |
| MyVoteSys <$\alpha$=0.50, $\beta$=0.33, $\gamma$=0.66, MaxVariance> | 1.225 | 0.833 | 0.000 |
| MyVoteSys <$\alpha$=0.50, $\beta$=____, $\gamma$=____, First> | 1.225 | 0.833 | 0.000 |
| MyVoteSys <$\alpha$=0.50, $\beta$=0.33, $\gamma$=____, MaxStDev> | 1.225 | 0.833 | 0.000 |
| MyVoteSys <$\alpha$=0.50, $\beta$=0.33, $\gamma$=0.80, MaxStDev> | 1.225 | 0.833 | 0.000 |
| MyVoteSys <$\alpha$=0.50, $\beta$=____, $\gamma$=0.80, MinEntropy> | 1.225 | 0.833 | 0.000 |
| MyVoteSys <α=0.50, β=0.33, γ=____, MaxVariance> | 1.225 | 0.833 | 0.000 |
| MyVoteSys <α=0.50, β=____, γ=0.80, MaxStDev> | 1.225 | 0.833 | 0.000 |
| PreferentialBlockVoting | 1.225 | 0.832 | 0.000 |
| SingleTransferableVote | 1.225 | 0.832 | 0.000 |
| MyVoteSys <α=____, β=____, γ=0.80, MinEntropy> | 1.225 | 0.832 | 0.000 |
| MyVoteSys <α=0.66, β=____, γ=0.80, MaxVariance> | 1.236 | 0.826 | 0.000 |
| MyVoteSys <α=0.66, β=____, γ=0.80, First> | 1.236 | 0.826 | 0.000 |
| MyVoteSys <α=0.66, β=____, γ=____, MinEntropy> | 1.236 | 0.826 | 0.000 |
| MyVoteSys <α=0.50, β=____, γ=0.66, Last> | 1.236 | 0.826 | 0.000 |
| MyVoteSys <α=0.66, β=____, γ=____, First> | 1.236 | 0.826 | 0.000 |
| MyVoteSys <α=0.50, β=____, γ=0.66, MaxEntropy> | 1.236 | 0.826 | 0.000 |
| MyVoteSys <α=0.66, β=____, γ=0.80, MaxStDev> | 1.236 | 0.826 | 0.000 |
| MyVoteSys <α=0.50, β=____, γ=0.66, MinVariance> | 1.236 | 0.826 | 0.000 |
| MyVoteSys <α=0.50, β=0.33, γ=0.66, Last> | 1.238 | 0.826 | 0.000 |
| MyVoteSys <α=0.50, β=0.33, γ=0.66, MaxEntropy> | 1.238 | 0.826 | 0.000 |
| MyVoteSys <α=0.50, β=0.33, γ=0.66, MinVariance> | 1.238 | 0.826 | 0.000 |
| MyVoteSys <α=0.66, β=0.33, γ=0.80, MaxVariance> | 1.250 | 0.824 | 0.000 |
| MyVoteSys <α=0.66, β=0.33, γ=____, MinEntropy> | 1.250 | 0.824 | 0.000 |
| MyVoteSys <α=0.66, β=0.33, γ=0.80, First> | 1.250 | 0.824 | 0.000 |
| MyVoteSys <α=0.66, β=0.33, γ=____, MaxVariance> | 1.250 | 0.824 | 0.000 |
| MyVoteSys <α=0.66, β=0.33, γ=0.80, MinEntropy> | 1.250 | 0.824 | 0.000 |
| MyVoteSys <α=0.66, β=0.33, γ=____, MaxStDev> | 1.250 | 0.824 | 0.000 |
| MyVoteSys <α=0.66, β=0.33, γ=____, First> | 1.250 | 0.824 | 0.000 |
| MyVoteSys <α=0.66, β=0.33, γ=0.80, MaxStDev> | 1.250 | 0.824 | 0.000 |
| MyVoteSys <α=0.50, β=0.33, γ=0.80, Last> | 1.288 | 0.792 | 0.000 |
| MyVoteSys <α=0.50, β=0.33, γ=0.80, MaxEntropy> | 1.288 | 0.792 | 0.000 |
| MyVoteSys <α=0.50, β=0.33, γ=0.80, MinVariance> | 1.288 | 0.792 | 0.000 |
| MyVoteSys <α=0.66, β=0.33, γ=0.80, MaxEntropy> | 1.300 | 0.789 | 0.000 |
| MyVoteSys <α=0.66, β=0.33, γ=0.80, Last> | 1.300 | 0.789 | 0.000 |
| MyVoteSys <α=0.66, β=0.33, γ=0.80, MinVariance> | 1.300 | 0.789 | 0.000 |
| MyVoteSys <α=0.50, β=____, γ=0.80, Last> | 1.308 | 0.789 | 0.000 |
| MyVoteSys <α=0.80, β=____, γ=____, MinEntropy> | 1.308 | 0.789 | 0.000 |
| MyVoteSys <α=0.80, β=____, γ=____, First> | 1.308 | 0.789 | 0.000 |
| MyVoteSys <α=0.66, β=____, γ=0.80, Last> | 1.308 | 0.789 | 0.000 |
| MyVoteSys <α=0.66, β=____, γ=0.80, MinVariance> | 1.308 | 0.789 | 0.000 |
| MyVoteSys <α=0.66, β=____, γ=0.80, MaxEntropy> | 1.308 | 0.789 | 0.000 |
| MyVoteSys <α=0.50, β=____, γ=0.80, MaxEntropy> | 1.308 | 0.789 | 0.000 |
| MyVoteSys <α=0.50, β=____, γ=0.80, MinVariance> | 1.308 | 0.789 | 0.000 |
| FirstPastThePost (but without tactical voting) | 1.325 | 0.773 | 0.000 |
| MyVoteSys <α=0.50, β=0.33, γ=____, MinVariance> | 1.337 | 0.756 | 0.000 |
| MyVoteSys <α=0.50, β=0.33, γ=____, Last> | 1.346 | 0.752 | 0.000 |
| MyVoteSys <α=0.50, β=0.33, γ=____, MaxEntropy> | 1.346 | 0.752 | 0.000 |
| MyVoteSys <α=0.66, β=0.33, γ=____, MinVariance> | 1.349 | 0.753 | 0.000 |
| MyVoteSys <α=0.66, β=0.33, γ=____, Last> | 1.358 | 0.749 | 0.000 |
| MyVoteSys <α=0.66, β=0.33, γ=____, MaxEntropy> | 1.358 | 0.749 | 0.000 |
| MyVoteSys <α=0.80, β=0.33, γ=____, MaxStDev> | 1.420 | 0.780 | 0.000 |
| MyVoteSys <α=0.80, β=0.33, γ=____, MaxVariance> | 1.420 | 0.780 | 0.000 |
| MyVoteSys <α=0.80, β=0.33, γ=____, MinEntropy> | 1.420 | 0.780 | 0.000 |
| MyVoteSys <α=0.80, β=0.33, γ=____, First> | 1.420 | 0.780 | 0.000 |
| MaxStandardDeviationWeightedSumWins<α=0> | 1.464 | 0.748 | 0.000 |
| MaxVarianceWeightedSumWins<α=0> | 1.465 | 0.745 | 0.000 |
| MyVoteSys <α=0.80, β=0.33, γ=____, MinVariance> | 1.469 | 0.744 | 0.000 |
| MaxStandardDeviationWeightedSumWins<α=0.5> | 1.470 | 0.748 | 0.000 |
| MaxVarianceWeightedSumWins<α=0.5> | 1.477 | 0.744 | 0.000 |
| MyVoteSys <α=0.80, β=0.33, γ=____, Last> | 1.478 | 0.740 | 0.000 |
| MyVoteSys <α=0.80, β=0.33, γ=____, MaxEntropy> | 1.478 | 0.740 | 0.000 |
| MyVoteSys <α=0.80, β=____, γ=____, MinVariance> | 1.504 | 0.730 | 0.000 |
| MyVoteSys <α=0.66, β=____, γ=____, MinVariance> | 1.504 | 0.730 | 0.000 |
| MyVoteSys <α=0.50, β=____, γ=____, MinVariance> | 1.504 | 0.730 | 0.000 |
| MaxVarianceProportionalWeightedSumWins<α=0> | 1.700 | 0.710 | 0.000 |
| MaxVarianceProportionalWeightedSumWins<α=0.5> | 1.710 | 0.709 | 0.000 |

| | meanWinnerRank | rateTrueWinners | rateWinner<NULL |
|---|---|---|---|
| MaxStandardDeviationProportionalWeightedSumWins<$\alpha$=0> | 1.852 | 0.694 | 0.000 |
| MaxStandardDeviationProportionalWeightedSumWin<$\alpha$=0.5>1.852 | | 0.694 | 0.000 |
| MyVoteSys <$\alpha$=0.50, $\beta$=____, $\gamma$=____, MaxStDev> | 2.444 | 0.644 | 0.000 |
| MyVoteSys <$\alpha$=0.50, $\beta$=____, $\gamma$=____, MaxVariance> | 2.444 | 0.644 | 0.000 |
| MyVoteSys <$\alpha$=0.66, $\beta$=____, $\gamma$=____, MaxVariance> | 2.457 | 0.642 | 0.000 |
| MyVoteSys <$\alpha$=0.66, $\beta$=____, $\gamma$=____, MaxStDev> | 2.457 | 0.642 | 0.000 |
| MyVoteSys <$\alpha$=0.50, $\beta$=____, $\gamma$=____, MaxEntropy> | 2.460 | 0.640 | 0.000 |
| MyVoteSys <$\alpha$=0.80, $\beta$=____, $\gamma$=____, Last> | 2.460 | 0.640 | 0.000 |
| MyVoteSys <$\alpha$=0.80, $\beta$=____, $\gamma$=____, MaxStDev> | 2.460 | 0.640 | 0.000 |
| MyVoteSys <$\alpha$=0.50, $\beta$=____, $\gamma$=____, Last> | 2.460 | 0.640 | 0.000 |
| MyVoteSys <$\alpha$=0.80, $\beta$=____, $\gamma$=____, MaxVariance> | 2.460 | 0.640 | 0.000 |
| MyVoteSys <$\alpha$=0.66, $\beta$=____, $\gamma$=____, MaxEntropy> | 2.460 | 0.640 | 0.000 |
| MyVoteSys <$\alpha$=0.80, $\beta$=____, $\gamma$=____, MaxEntropy> | 2.460 | 0.640 | 0.000 |
| MyVoteSys <$\alpha$=0.66, $\beta$=____, $\gamma$=____, Last> | 2.460 | 0.640 | 0.000 |
| bestVoter (Dictatorship) | 2.594 | 0.417 | 0.010 |
| MinEntropyProportionalWeightedSumWins<$\alpha$=0> | 3.944 | 0.141 | 0.050 |
| MinEntropyWeightedSumWins2<$\alpha$=0> | 3.944 | 0.141 | 0.050 |
| MinEntropyProportionalWeightedSumWins2<$\alpha$=0> | 3.944 | 0.141 | 0.050 |
| MinEntropyWeightedSumWins<$\alpha$=0> | 3.944 | 0.141 | 0.050 |
| MinEntropyWeightedSumWins2<$\alpha$=0.5> | 3.949 | 0.141 | 0.050 |
| MinEntropyWeightedSumWins<$\alpha$=0.5> | 3.949 | 0.141 | 0.050 |
| MinEntropyProportionalWeightedSumWins<$\alpha$=0.5> | 3.949 | 0.141 | 0.050 |
| MinEntropyProportionalWeightedSumWins2<$\alpha$=0.5> | 3.949 | 0.141 | 0.050 |

| Metrics | val_MeanSquaredErr |
|---|---|
| Algorithms | |
| crowd-Mean | 21834.503906 |
| crowd-Median | 21684.726562 |
| bestVoter (Dictatorship) | 7147.884766 |

================================ END OF SIMULATION 5 ================================

================================ START OF SIMULATION 6 ================================

numCandiates : 20
numVoters : 750
numElections : 750
columnBlindness : 1
epochs : 166
trainableLayerCount : 1
crowdBuildMethod : {'name': 'standardDistribution', 'mean': 10000, 'standardDeviation': 2500}
dataSetName : mySynthetic
predictedFeature : y

======== SIMULATION RESULTS ========

| Metrics | meanWinnerRank | rateTrueWinners | rateWinner<NULL |
|---|---|---|---|
| Algorithms | | | |
| MyVoteSys <$\alpha$=0.50, $\beta$=____, $\gamma$=0.66, MinEntropy> | 1.102 | 0.917 | 0.000 |
| MyVoteSys <$\alpha$=0.50, $\beta$=____, $\gamma$=0.80, First> | 1.102 | 0.917 | 0.000 |
| MyVoteSys <$\alpha$=0.50, $\beta$=0.33, $\gamma$=0.66, MinEntropy> | 1.102 | 0.917 | 0.000 |
| MyVoteSys <$\alpha$=0.50, $\beta$=0.33, $\gamma$=____, MaxVariance> | 1.102 | 0.917 | 0.000 |
| MyVoteSys <$\alpha$=0.50, $\beta$=0.33, $\gamma$=0.80, MaxVariance> | 1.102 | 0.917 | 0.000 |
| MyVoteSys <$\alpha$=0.50, $\beta$=____, $\gamma$=0.80, MaxVariance> | 1.102 | 0.917 | 0.000 |
| MyVoteSys <$\alpha$=0.50, $\beta$=____, $\gamma$=0.80, MinEntropy> | 1.102 | 0.917 | 0.000 |
| MyVoteSys <$\alpha$=0.50, $\beta$=____, $\gamma$=0.66, First> | 1.102 | 0.917 | 0.000 |
| MyVoteSys <$\alpha$=0.50, $\beta$=0.33, $\gamma$=0.66, MaxVariance> | 1.102 | 0.917 | 0.000 |
| MyVoteSys <$\alpha$=0.50, $\beta$=0.33, $\gamma$=0.66, First> | 1.102 | 0.917 | 0.000 |
| MyVoteSys <$\alpha$=0.50, $\beta$=____, $\gamma$=0.66, MaxVariance> | 1.102 | 0.917 | 0.000 |
| MyVoteSys <$\alpha$=0.50, $\beta$=____, $\gamma$=____, First> | 1.102 | 0.917 | 0.000 |
| MyVoteSys <$\alpha$=0.50, $\beta$=0.33, $\gamma$=____, MinEntropy> | 1.102 | 0.917 | 0.000 |
| MyVoteSys <$\alpha$=0.50, $\beta$=____, $\gamma$=____, MinEntropy> | 1.102 | 0.917 | 0.000 |
| MyVoteSys <$\alpha$=0.50, $\beta$=0.33, $\gamma$=0.80, MinEntropy> | 1.102 | 0.917 | 0.000 |
| MyVoteSys <$\alpha$=0.50, $\beta$=0.33, $\gamma$=0.80, First> | 1.102 | 0.917 | 0.000 |
| MyVoteSys <α=0.50, β=0.33, γ=____, First> | 1.102 | 0.917 | 0.000 |
| InstantRunoffVoting | 1.109 | 0.920 | 0.000 |
| crowd−Mean | 1.109 | 0.918 | 0.000 |
| crowd−Median | 1.109 | 0.918 | 0.000 |
| PreferentialBlockVoting | 1.110 | 0.918 | 0.000 |
| SingleTransferableVote | 1.110 | 0.918 | 0.000 |
| MyVoteSys <α=0.50, β=0.33, γ=0.66, MaxEntropy> | 1.137 | 0.890 | 0.000 |
| MyVoteSys <α=0.50, β=____, γ=0.66, MinVariance> | 1.137 | 0.890 | 0.000 |
| MyVoteSys <α=0.50, β=____, γ=0.66, Last> | 1.137 | 0.890 | 0.000 |
| MyVoteSys <α=0.66, β=0.33, γ=0.80, MinEntropy> | 1.137 | 0.890 | 0.000 |
| MyVoteSys <α=0.66, β=0.33, γ=____, MaxVariance> | 1.137 | 0.890 | 0.000 |
| MyVoteSys <α=0.66, β=0.33, γ=0.80, MaxVariance> | 1.137 | 0.890 | 0.000 |
| MyVoteSys <α=0.66, β=____, γ=____, First> | 1.137 | 0.890 | 0.000 |
| MyVoteSys <α=0.66, β=0.33, γ=____, MinEntropy> | 1.137 | 0.890 | 0.000 |
| MyVoteSys <α=0.50, β=0.33, γ=0.66, MinVariance> | 1.137 | 0.890 | 0.000 |
| MyVoteSys <α=0.66, β=____, γ=0.80, MaxVariance> | 1.137 | 0.890 | 0.000 |
| MyVoteSys <α=0.66, β=____, γ=____, MinEntropy> | 1.137 | 0.890 | 0.000 |
| MyVoteSys <α=0.66, β=____, γ=0.80, First> | 1.137 | 0.890 | 0.000 |
| MyVoteSys <α=0.66, β=0.33, γ=0.80, First> | 1.137 | 0.890 | 0.000 |
| MyVoteSys <α=0.50, β=____, γ=0.66, MaxEntropy> | 1.137 | 0.890 | 0.000 |
| MyVoteSys <α=0.50, β=0.33, γ=0.66, Last> | 1.137 | 0.890 | 0.000 |
| MyVoteSys <α=0.66, β=____, γ=0.80, MinEntropy> | 1.137 | 0.890 | 0.000 |
| MyVoteSys <α=0.66, β=0.33, γ=____, First> | 1.137 | 0.890 | 0.000 |
| FirstPastThePost (but without tactical voting) | 1.140 | 0.905 | 0.000 |
| MyVoteSys <α=0.66, β=____, γ=0.80, Last> | 1.154 | 0.873 | 0.000 |
| MyVoteSys <α=0.80, β=____, γ=____, First> | 1.154 | 0.873 | 0.000 |
| MyVoteSys <α=0.80, β=____, γ=____, MinEntropy> | 1.154 | 0.873 | 0.000 |
| MyVoteSys <α=0.66, β=____, γ=0.80, MaxEntropy> | 1.154 | 0.873 | 0.000 |
| MyVoteSys <α=0.66, β=0.33, γ=0.80, Last> | 1.154 | 0.873 | 0.000 |
| MyVoteSys <α=0.66, β=0.33, γ=0.80, MinVariance> | 1.154 | 0.873 | 0.000 |
| MyVoteSys <α=0.66, β=0.33, γ=0.80, MaxEntropy> | 1.154 | 0.873 | 0.000 |
| MyVoteSys <α=0.50, β=0.33, γ=0.80, MinVariance> | 1.154 | 0.873 | 0.000 |
| MyVoteSys <α=0.50, β=0.33, γ=0.80, Last> | 1.154 | 0.873 | 0.000 |
| MyVoteSys <α=0.66, β=____, γ=0.80, MinVariance> | 1.154 | 0.873 | 0.000 |
| MyVoteSys <α=0.50, β=0.33, γ=0.80, MaxEntropy> | 1.154 | 0.873 | 0.000 |
| MyVoteSys <α=0.50, β=____, γ=0.80, Last> | 1.154 | 0.873 | 0.000 |
| MyVoteSys <α=0.50, β=____, γ=0.80, MaxEntropy> | 1.154 | 0.873 | 0.000 |
| MyVoteSys <α=0.50, β=0.33, γ=0.80, MinVariance> | 1.154 | 0.873 | 0.000 |
| MyVoteSys <α=0.80, β=0.33, γ=____, MinEntropy> | 1.169 | 0.872 | 0.000 |
| MyVoteSys <α=0.80, β=0.33, γ=____, First> | 1.169 | 0.872 | 0.000 |
| MyVoteSys <α=0.80, β=0.33, γ=____, MaxVariance> | 1.169 | 0.872 | 0.000 |
| MyVoteSys <α=0.50, β=0.33, γ=____, MinVariance> | 1.473 | 0.680 | 0.000 |
| MyVoteSys <α=0.66, β=0.33, γ=____, MinVariance> | 1.473 | 0.680 | 0.000 |
| bestVoter (Dictatorship) | 1.478 | 0.698 | 0.000 |
| MyVoteSys <α=0.80, β=0.33, γ=____, MinVariance> | 1.488 | 0.678 | 0.000 |
| MyVoteSys <α=0.66, β=____, γ=____, MinVariance> | 1.572 | 0.637 | 0.000 |
| MyVoteSys <α=0.50, β=____, γ=____, MinVariance> | 1.572 | 0.637 | 0.000 |
| MyVoteSys <α=0.80, β=____, γ=____, MinVariance> | 1.572 | 0.637 | 0.000 |
| MyVoteSys <α=0.66, β=0.33, γ=____, MaxEntropy> | 1.618 | 0.626 | 0.000 |
| MyVoteSys <α=0.50, β=0.33, γ=____, MaxEntropy> | 1.618 | 0.626 | 0.000 |
| MyVoteSys <α=0.66, β=0.33, γ=____, Last> | 1.618 | 0.626 | 0.000 |
| MyVoteSys <α=0.50, β=0.33, γ=____, Last> | 1.618 | 0.626 | 0.000 |
| MyVoteSys <α=0.80, β=0.33, γ=____, Last> | 1.633 | 0.625 | 0.000 |
| MyVoteSys <α=0.80, β=0.33, γ=____, MaxEntropy> | 1.633 | 0.625 | 0.000 |
| MyVoteSys <α=0.50, β=____, γ=____, MaxVariance> | 2.037 | 0.545 | 0.004 |
| MyVoteSys <α=0.66, β=____, γ=____, MaxVariance> | 2.238 | 0.469 | 0.005 |
| MyVoteSys <α=0.80, β=____, γ=____, MaxVariance> | 2.362 | 0.422 | 0.005 |
| MyVoteSys <α=0.66, β=____, γ=____, Last> | 2.381 | 0.413 | 0.005 |
| MyVoteSys <α=0.50, β=____, γ=____, MaxEntropy> | 2.381 | 0.413 | 0.005 |
| MyVoteSys <α=0.80, β=____, γ=____, Last> | 2.381 | 0.413 | 0.005 |
| MyVoteSys <α=0.50, β=____, γ=____, Last> | 2.381 | 0.413 | 0.005 |
| MyVoteSys <$\alpha$=0.80, $\beta$=____, $\gamma$=____, MaxEntropy> | 2.381 | 0.413 | 0.005 |
| MyVoteSys <$\alpha$=0.66, $\beta$=____, $\gamma$=____, MaxEntropy> | 2.381 | 0.413 | 0.005 |

| Metrics | val_MeanSquaredErr |
|---|---|
| Algorithms | |
| crowd-Mean | 18234.189453 |
| crowd-Median | 17967.429688 |
| bestVoter (Dictatorship) | 3134.116699 |

============================= END OF SIMULATION 6 =================================

============================= START OF SIMULATION 7 =================================

numCandiates : 20
numVoters : 750
numElections : 750
columnBlindness : 1
epochs : 166
trainableLayerCount : 1
crowdBuildMethod : {'name': 'standardDistribution', 'mean': 10000, 'standardDeviation': 2500}
dataSetName : mySynthetic
predictedFeature : y

======== SIMULATION RESULTS ========

| Metrics | meanWinnerRank | rateTrueWinners | rateWinner<NULL |
|---|---|---|---|
| Algorithms | | | |
| MyVoteSys <$\alpha$=0.50, $\beta$=0.33, $\gamma$=0.80, First> | 1.129 | 0.890 | 0.0 |
| MyVoteSys <$\alpha$=0.50, $\beta$=____, $\gamma$=____, First> | 1.129 | 0.890 | 0.0 |
| MyVoteSys <$\alpha$=0.50, $\beta$=____, $\gamma$=0.66, MinEntropy> | 1.129 | 0.890 | 0.0 |
| MyVoteSys <$\alpha$=0.50, $\beta$=0.33, $\gamma$=0.66, MinEntropy> | 1.129 | 0.890 | 0.0 |
| MyVoteSys <$\alpha$=0.50, $\beta$=____, $\gamma$=0.66, MaxVariance> | 1.129 | 0.890 | 0.0 |
| MyVoteSys <$\alpha$=0.50, $\beta$=0.33, $\gamma$=0.80, MaxStDev> | 1.129 | 0.890 | 0.0 |
| MyVoteSys <$\alpha$=0.50, $\beta$=____, $\gamma$=0.80, MaxVariance> | 1.129 | 0.890 | 0.0 |
| MyVoteSys <$\alpha$=0.50, $\beta$=0.33, $\gamma$=____, First> | 1.129 | 0.890 | 0.0 |
| MyVoteSys <$\alpha$=0.50, $\beta$=____, $\gamma$=0.80, MinEntropy> | 1.129 | 0.890 | 0.0 |
| MyVoteSys <$\alpha$=0.50, $\beta$=0.33, $\gamma$=____, MaxStDev> | 1.129 | 0.890 | 0.0 |
| MyVoteSys <$\alpha$=0.50, $\beta$=____, $\gamma$=0.80, MaxStDev> | 1.129 | 0.890 | 0.0 |
| MyVoteSys <$\alpha$=0.50, $\beta$=0.33, $\gamma$=0.80, MinEntropy> | 1.129 | 0.890 | 0.0 |
| MyVoteSys <$\alpha$=0.50, $\beta$=0.33, $\gamma$=0.66, MaxVariance> | 1.129 | 0.890 | 0.0 |
| MyVoteSys <$\alpha$=0.50, $\beta$=____, $\gamma$=0.66, MaxStDev> | 1.129 | 0.890 | 0.0 |
| MyVoteSys <$\alpha$=0.50, $\beta$=0.33, $\gamma$=____, MaxVariance> | 1.129 | 0.890 | 0.0 |
| MyVoteSys <$\alpha$=0.50, $\beta$=____, $\gamma$=____, MinEntropy> | 1.129 | 0.890 | 0.0 |
| MyVoteSys <$\alpha$=0.50, $\beta$=____, $\gamma$=0.80, First> | 1.129 | 0.890 | 0.0 |
| MyVoteSys <$\alpha$=0.50, $\beta$=0.33, $\gamma$=____, MinEntropy> | 1.129 | 0.890 | 0.0 |
| MyVoteSys <$\alpha$=0.50, $\beta$=0.33, $\gamma$=0.66, First> | 1.129 | 0.890 | 0.0 |
| MyVoteSys <$\alpha$=0.50, $\beta$=____, $\gamma$=0.66, First> | 1.129 | 0.890 | 0.0 |
| MyVoteSys <$\alpha$=0.50, $\beta$=0.33, $\gamma$=0.80, MaxVariance> | 1.129 | 0.890 | 0.0 |
| MyVoteSys <$\alpha$=0.50, $\beta$=0.33, $\gamma$=0.66, MaxStDev> | 1.129 | 0.890 | 0.0 |
| PreferentialBlockVoting | 1.134 | 0.885 | 0.0 |
| SingleTransferableVote | 1.134 | 0.885 | 0.0 |
| InstantRunoffVoting | 1.134 | 0.885 | 0.0 |
| crowd-Median | 1.136 | 0.885 | 0.0 |
| crowd-Mean | 1.136 | 0.885 | 0.0 |
| FirstPastThePost (but without tactical voting) | 1.149 | 0.874 | 0.0 |
| MyVoteSys <$\alpha$=0.50, $\beta$=0.33, $\gamma$=0.80, Last> | 1.156 | 0.862 | 0.0 |
| MyVoteSys <$\alpha$=0.50, $\beta$=0.33, $\gamma$=0.80, MaxEntropy> | 1.156 | 0.862 | 0.0 |
| MyVoteSys <$\alpha$=0.80, $\beta$=0.33, $\gamma$=____, MinEntropy> | 1.156 | 0.862 | 0.0 |
| MyVoteSys <$\alpha$=0.50, $\beta$=____, $\gamma$=0.80, MinVariance> | 1.156 | 0.862 | 0.0 |
| MyVoteSys <$\alpha$=0.66, $\beta$=____, $\gamma$=0.80, Last> | 1.156 | 0.862 | 0.0 |
| MyVoteSys <$\alpha$=0.80, $\beta$=0.33, $\gamma$=____, First> | 1.156 | 0.862 | 0.0 |
| MyVoteSys <$\alpha$=0.66, $\beta$=0.33, $\gamma$=0.80, MaxEntropy> | 1.156 | 0.862 | 0.0 |
| MyVoteSys <$\alpha$=0.50, $\beta$=____, $\gamma$=0.80, Last> | 1.156 | 0.862 | 0.0 |
| MyVoteSys <$\alpha$=0.50, $\beta$=____, $\gamma$=0.80, MaxEntropy> | 1.156 | 0.862 | 0.0 |
| MyVoteSys ⟨α=0.80, β=____, γ=____, First⟩ | 1.156 | 0.862 | 0.0 |
| MyVoteSys ⟨α=0.66, β=0.33, γ=0.80, Last⟩ | 1.156 | 0.862 | 0.0 |
| MyVoteSys ⟨α=0.50, β=0.33, γ=0.80, MinVariance⟩ | 1.156 | 0.862 | 0.0 |
| MyVoteSys ⟨α=0.66, β=____, γ=0.80, MaxEntropy⟩ | 1.156 | 0.862 | 0.0 |
| MyVoteSys ⟨α=0.66, β=____, γ=0.80, MinVariance⟩ | 1.156 | 0.862 | 0.0 |
| MyVoteSys ⟨α=0.66, β=0.33, γ=0.80, MinVariance⟩ | 1.156 | 0.862 | 0.0 |
| MyVoteSys ⟨α=0.80, β=____, γ=____, MinEntropy⟩ | 1.156 | 0.862 | 0.0 |
| MyVoteSys ⟨α=0.66, β=0.33, γ=0.66, MinVariance⟩ | 1.157 | 0.865 | 0.0 |
| MyVoteSys ⟨α=0.50, β=____, γ=0.66, MaxEntropy⟩ | 1.157 | 0.865 | 0.0 |
| MyVoteSys ⟨α=0.66, β=0.33, γ=0.80, MaxVariance⟩ | 1.157 | 0.865 | 0.0 |
| MyVoteSys ⟨α=0.66, β=0.33, γ=____, MaxVariance⟩ | 1.157 | 0.865 | 0.0 |
| MyVoteSys ⟨α=0.66, β=0.33, γ=____, MinEntropy⟩ | 1.157 | 0.865 | 0.0 |
| MyVoteSys ⟨α=0.50, β=0.33, γ=0.66, MaxEntropy⟩ | 1.157 | 0.865 | 0.0 |
| MyVoteSys ⟨α=0.66, β=0.33, γ=0.80, MaxStDev⟩ | 1.157 | 0.865 | 0.0 |
| MyVoteSys ⟨α=0.50, β=0.33, γ=0.66, Last⟩ | 1.157 | 0.865 | 0.0 |
| MyVoteSys ⟨α=0.66, β=0.33, γ=____, First⟩ | 1.157 | 0.865 | 0.0 |
| MyVoteSys ⟨α=0.66, β=____, γ=0.80, MaxStDev⟩ | 1.157 | 0.865 | 0.0 |
| MyVoteSys ⟨α=0.66, β=____, γ=0.80, First⟩ | 1.157 | 0.865 | 0.0 |
| MyVoteSys ⟨α=0.66, β=____, γ=0.80, MinEntropy⟩ | 1.157 | 0.865 | 0.0 |
| MyVoteSys ⟨α=0.66, β=____, γ=____, First⟩ | 1.157 | 0.865 | 0.0 |
| MyVoteSys ⟨α=0.66, β=0.33, γ=0.80, First⟩ | 1.157 | 0.865 | 0.0 |
| MyVoteSys ⟨α=0.50, β=____, γ=0.66, MinVariance⟩ | 1.157 | 0.865 | 0.0 |
| MyVoteSys ⟨α=0.50, β=____, γ=0.66, Last⟩ | 1.157 | 0.865 | 0.0 |
| MyVoteSys ⟨α=0.66, β=____, γ=____, MinEntropy⟩ | 1.157 | 0.865 | 0.0 |
| MyVoteSys ⟨α=0.66, β=0.33, γ=0.80, MinEntropy⟩ | 1.157 | 0.865 | 0.0 |
| MyVoteSys ⟨α=0.66, β=0.33, γ=____, MaxStDev⟩ | 1.157 | 0.865 | 0.0 |
| MyVoteSys ⟨α=0.66, β=____, γ=0.80, MaxVariance⟩ | 1.157 | 0.865 | 0.0 |
| MyVoteSys ⟨α=0.80, β=0.33, γ=____, MaxVariance⟩ | 1.158 | 0.861 | 0.0 |
| MyVoteSys ⟨α=0.80, β=0.33, γ=____, MaxStDev⟩ | 1.158 | 0.861 | 0.0 |
| MyVoteSys ⟨α=0.50, β=____, γ=____, MaxStDev⟩ | 1.273 | 0.796 | 0.0 |
| MyVoteSys ⟨α=0.50, β=____, γ=____, MaxVariance⟩ | 1.273 | 0.796 | 0.0 |
| bestVoter (Dictatorship) | 1.337 | 0.762 | 0.0 |
| MyVoteSys ⟨α=0.66, β=____, γ=____, MaxVariance⟩ | 1.357 | 0.737 | 0.0 |
| MyVoteSys ⟨α=0.66, β=____, γ=____, MaxStDev⟩ | 1.357 | 0.737 | 0.0 |
| MyVoteSys ⟨α=0.80, β=0.33, γ=____, MinVariance⟩ | 1.378 | 0.720 | 0.0 |
| MyVoteSys ⟨α=0.66, β=0.33, γ=____, MinVariance⟩ | 1.378 | 0.720 | 0.0 |
| MyVoteSys ⟨α=0.50, β=0.33, γ=____, MinVariance⟩ | 1.378 | 0.720 | 0.0 |
| MyVoteSys ⟨α=0.80, β=____, γ=____, MaxVariance⟩ | 1.381 | 0.720 | 0.0 |
| MyVoteSys ⟨α=0.80, β=____, γ=____, MaxStDev⟩ | 1.381 | 0.720 | 0.0 |
| MyVoteSys ⟨α=0.80, β=____, γ=____, MinVariance⟩ | 1.386 | 0.718 | 0.0 |
| MyVoteSys ⟨α=0.66, β=____, γ=____, MinVariance⟩ | 1.386 | 0.718 | 0.0 |
| MyVoteSys ⟨α=0.50, β=____, γ=____, MinVariance⟩ | 1.386 | 0.718 | 0.0 |
| MyVoteSys ⟨α=0.66, β=0.33, γ=____, Last⟩ | 1.388 | 0.714 | 0.0 |
| MyVoteSys ⟨α=0.50, β=0.33, γ=____, MaxEntropy⟩ | 1.388 | 0.714 | 0.0 |
| MyVoteSys ⟨α=0.80, β=0.33, γ=____, Last⟩ | 1.388 | 0.714 | 0.0 |
| MyVoteSys ⟨α=0.80, β=0.33, γ=____, MaxEntropy⟩ | 1.388 | 0.714 | 0.0 |
| MyVoteSys ⟨α=0.66, β=0.33, γ=____, MaxEntropy⟩ | 1.388 | 0.714 | 0.0 |
| MyVoteSys ⟨α=0.66, β=0.33, γ=____, Last⟩ | 1.388 | 0.714 | 0.0 |
| MaxStandardDeviationWeightedSumWins⟨α=0.5⟩ | 1.393 | 0.713 | 0.0 |
| MaxStandardDeviationProportionalWeightedSumWins⟨α=0⟩ | 1.393 | 0.713 | 0.0 |
| MyVoteSys ⟨α=0.50, β=____, γ=____, Last⟩ | 1.393 | 0.713 | 0.0 |
| MaxStandardDeviationProportionalWeightedSumWins⟨α=0.5⟩ | 1.393 | 0.713 | 0.0 |
| MyVoteSys ⟨α=0.66, β=____, γ=____, MaxEntropy⟩ | 1.393 | 0.713 | 0.0 |
| MyVoteSys ⟨α=0.80, β=____, γ=____, Last⟩ | 1.393 | 0.713 | 0.0 |
| MaxVarianceProportionalWeightedSumWins⟨α=0.5⟩ | 1.393 | 0.713 | 0.0 |
| MaxVarianceProportionalWeightedSumWins⟨α=0⟩ | 1.393 | 0.713 | 0.0 |
| MyVoteSys ⟨α=0.80, β=____, γ=____, MaxEntropy⟩ | 1.393 | 0.713 | 0.0 |
| MyVoteSys ⟨α=0.50, β=____, γ=____, MaxEntropy⟩ | 1.393 | 0.713 | 0.0 |
| MyVoteSys ⟨α=0.66, β=____, γ=____, Last⟩ | 1.393 | 0.713 | 0.0 |
| MaxVarianceWeightedSumWins⟨α=0.5⟩ | 1.394 | 0.713 | 0.0 |
| MaxVarianceWeightedSumWins⟨α=0⟩ | 1.394 | 0.713 | 0.0 |
| MaxStandardDeviationWeightedSumWins<$\alpha$=0> | 1.394 | 0.713 | 0.0 |
| MinEntropyWeightedSumWins2<$\alpha$=0> | 2.493 | 0.212 | 0.0 |
| MinEntropyProportionalWeightedSumWins2<$\alpha$=0> | 2.493 | 0.212 | 0.0 |
| MinEntropyWeightedSumWins2<$\alpha$=0.5> | 2.493 | 0.212 | 0.0 |
| MinEntropyProportionalWeightedSumWins <$\alpha$=0> | 2.493 | 0.212 | 0.0 |
| MinEntropyProportionalWeightedSumWins <$\alpha$=0.5> | 2.493 | 0.212 | 0.0 |
| MinEntropyWeightedSumWins<$\alpha$=0> | 2.493 | 0.212 | 0.0 |
| MinEntropyWeightedSumWins<$\alpha$=0.5> | 2.493 | 0.212 | 0.0 |
| MinEntropyProportionalWeightedSumWins2<$\alpha$=0.5> | 2.493 | 0.212 | 0.0 |

| Metrics | val_MeanSquaredErr |
|---|---|
| Algorithms | |
| bestVoter (Dictatorship) | 2424.916992 |
| crowd-Median | 22154.931641 |
| crowd-Mean | 22559.896484 |

================================ END OF SIMULATION 7 ================================

================================ START OF SIMULATION 8 ================================

numCandiates : 20
numVoters : 750
numElections : 750
columnBlindness : [7, 9]
epochs : 133
trainableLayerCount : 1
crowdBuildMethod : {'name': 'standardDistribution', 'mean': 14000, 'standardDeviation': 1333}
dataSetName : mySynthetic
predictedFeature : y
========= SIMULATION RESULTS =======

| Metrics | meanWinnerRank | rateTrueWinners | rateWinner<NULL |
|---|---|---|---|
| Algorithms | | | |
| crowd-Median | 1.174 | 0.862 | 0.000 |
| crowd-Mean | 1.174 | 0.862 | 0.000 |
| MyVoteSys <$\alpha$=0.50, $\beta$=____, $\gamma$=0.66, MaxVariance> | 1.265 | 0.809 | 0.000 |
| MyVoteSys <$\alpha$=0.50, $\beta$=____, $\gamma$=0.66, MaxStDev> | 1.265 | 0.809 | 0.000 |
| MyVoteSys <$\alpha$=0.50, $\beta$=0.33, $\gamma$=0.66, MinEntropy> | 1.266 | 0.808 | 0.000 |
| MyVoteSys <$\alpha$=0.50, $\beta$=____, $\gamma$=0.80, MinEntropy> | 1.266 | 0.808 | 0.000 |
| MyVoteSys <$\alpha$=0.50, $\beta$=____, $\gamma$=____, First > | 1.266 | 0.808 | 0.000 |
| MyVoteSys <$\alpha$=0.50, $\beta$=____, $\gamma$=0.80, First > | 1.266 | 0.808 | 0.000 |
| MyVoteSys <$\alpha$=0.50, $\beta$=0.33, $\gamma$=____, MaxVariance> | 1.266 | 0.808 | 0.000 |
| MyVoteSys <$\alpha$=0.50, $\beta$=0.33, $\gamma$=0.80, MinEntropy> | 1.266 | 0.808 | 0.000 |
| MyVoteSys <$\alpha$=0.50, $\beta$=____, $\gamma$=0.66, First > | 1.266 | 0.808 | 0.000 |
| MyVoteSys <$\alpha$=0.50, $\beta$=____, $\gamma$=____, MinEntropy> | 1.266 | 0.808 | 0.000 |
| MyVoteSys <$\alpha$=0.50, $\beta$=0.33, $\gamma$=0.80, MaxStDev> | 1.266 | 0.808 | 0.000 |
| MyVoteSys <$\alpha$=0.50, $\beta$=0.33, $\gamma$=0.80, MaxVariance> | 1.266 | 0.808 | 0.000 |
| MyVoteSys <$\alpha$=0.50, $\beta$=0.33, $\gamma$=____, First > | 1.266 | 0.808 | 0.000 |
| MyVoteSys <$\alpha$=0.50, $\beta$=0.33, $\gamma$=0.66, MaxStDev> | 1.266 | 0.808 | 0.000 |
| MyVoteSys <$\alpha$=0.50, $\beta$=0.33, $\gamma$=0.80, First > | 1.266 | 0.808 | 0.000 |
| MyVoteSys <$\alpha$=0.50, $\beta$=0.33, $\gamma$=____, MaxStDev> | 1.266 | 0.808 | 0.000 |
| MyVoteSys <$\alpha$=0.50, $\beta$=____, $\gamma$=0.66, MinEntropy> | 1.266 | 0.808 | 0.000 |
| MyVoteSys <$\alpha$=0.50, $\beta$=0.33, $\gamma$=0.66, First > | 1.266 | 0.808 | 0.000 |
| MyVoteSys <$\alpha$=0.50, $\beta$=0.33, $\gamma$=0.66, MaxVariance> | 1.266 | 0.808 | 0.000 |
| MyVoteSys <$\alpha$=0.50, $\beta$=0.33, $\gamma$=____, MinEntropy> | 1.266 | 0.808 | 0.000 |
| MyVoteSys <$\alpha$=0.50, $\beta$=____, $\gamma$=0.80, MaxVariance> | 1.280 | 0.804 | 0.000 |
| MyVoteSys <$\alpha$=0.50, $\beta$=____, $\gamma$=0.80, MaxStDev> | 1.280 | 0.804 | 0.000 |
| MyVoteSys <$\alpha$=0.50, $\beta$=0.33, $\gamma$=0.66, Last> | 1.304 | 0.772 | 0.000 |
| MyVoteSys <$\alpha$=0.50, $\beta$=0.33, $\gamma$=0.66, MaxEntropy> | 1.304 | 0.772 | 0.000 |
| MyVoteSys <$\alpha$=0.50, $\beta$=____, $\gamma$=0.66, MaxEntropy> | 1.305 | 0.772 | 0.000 |
| MyVoteSys <$\alpha$=0.66, $\beta$=____, $\gamma$=____, First > | 1.305 | 0.772 | 0.000 |
| MyVoteSys <$\alpha$=0.50, $\beta$=____, $\gamma$=0.66, Last> | 1.305 | 0.772 | 0.000 |
| MyVoteSys <$\alpha$=0.66, $\beta$=____, $\gamma$=0.80, First > | 1.305 | 0.772 | 0.000 |
| MyVoteSys <$\alpha$=0.66, $\beta$=____, $\gamma$=____, MinEntropy> | 1.305 | 0.772 | 0.000 |
| MyVoteSys <$\alpha$=0.66, $\beta$=____, $\gamma$=0.80, MinEntropy> | 1.305 | 0.772 | 0.000 |
| MyVoteSys <$\alpha$=0.50, $\beta$=0.33, $\gamma$=0.66, MinVariance> | 1.305 | 0.772 | 0.000 |
| MyVoteSys <$\alpha$=0.50, $\beta$=____, $\gamma$=0.66, MinVariance> | 1.305 | 0.772 | 0.000 |
| MyVoteSys <$\alpha$=0.66, $\beta$=0.33, $\gamma$=0.80, MinEntropy> | 1.356 | 0.768 | 0.000 |
| MyVoteSys <$\alpha$=0.66, $\beta$=0.33, $\gamma$=0.80, First> | 1.356 | 0.768 | 0.000 |
| MyVoteSys <$\alpha$=0.66, $\beta$=0.33, $\gamma$=____, First> | 1.356 | 0.768 | 0.000 |
| MyVoteSys <$\alpha$=0.66, $\beta$=0.33, $\gamma$=____, MinEntropy> | 1.356 | 0.768 | 0.000 |
| MyVoteSys <$\alpha$=0.66, $\beta$=0.33, $\gamma$=0.80, MaxStDev> | 1.356 | 0.764 | 0.000 |
| MyVoteSys <$\alpha$=0.66, $\beta$=0.33, $\gamma$=0.80, MaxVariance> | 1.356 | 0.764 | 0.000 |
| MyVoteSys <$\alpha$=0.66, $\beta$=____, $\gamma$=0.80, MaxStDev> | 1.356 | 0.756 | 0.000 |
| MyVoteSys <$\alpha$=0.66, $\beta$=____, $\gamma$=0.80, MaxVariance> | 1.356 | 0.756 | 0.000 |
| MyVoteSys <$\alpha$=0.66, $\beta$=0.33, $\gamma$=____, MaxVariance> | 1.361 | 0.764 | 0.000 |
| MyVoteSys <$\alpha$=0.66, $\beta$=0.33, $\gamma$=____, MaxStDev> | 1.361 | 0.764 | 0.000 |
| InstantRunoffVoting | 1.412 | 0.753 | 0.000 |
| SingleTransferableVote | 1.412 | 0.750 | 0.000 |
| PreferentialBlockVoting | 1.412 | 0.750 | 0.000 |
| MyVoteSys <$\alpha$=0.50, $\beta$=0.33, $\gamma$=0.80, MinVariance> | 1.420 | 0.710 | 0.000 |
| MaxVarianceWeightedSumWins<$\alpha$=0> | 1.429 | 0.708 | 0.000 |
| MyVoteSys <$\alpha$=0.50, $\beta$=____, $\gamma$=0.80, MinVariance> | 1.429 | 0.706 | 0.000 |
| MaxStandardDeviationWeightedSumWins<$\alpha$=0> | 1.430 | 0.710 | 0.000 |
| MyVoteSys <$\alpha$=0.50, $\beta$=0.33, $\gamma$=____, MinVariance> | 1.430 | 0.704 | 0.000 |
| MyVoteSys <$\alpha$=0.50, $\beta$=0.33, $\gamma$=0.80, Last> | 1.430 | 0.701 | 0.000 |
| MyVoteSys <$\alpha$=0.50, $\beta$=0.33, $\gamma$=0.80, MaxEntropy> | 1.430 | 0.701 | 0.000 |
| MyVoteSys <$\alpha$=0.66, $\beta$=____, $\gamma$=0.80, MinVariance> | 1.442 | 0.705 | 0.000 |
| MyVoteSys <$\alpha$=0.50, $\beta$=0.33, $\gamma$=____, Last> | 1.449 | 0.688 | 0.000 |
| MyVoteSys <$\alpha$=0.50, $\beta$=0.33, $\gamma$=____, MaxEntropy> | 1.449 | 0.688 | 0.000 |
| MyVoteSys <$\alpha$=0.50, $\beta$=____, $\gamma$=____, MinVariance> | 1.460 | 0.696 | 0.000 |
| MaxVarianceWeightedSumWins<$\alpha$=0.5> | 1.462 | 0.693 | 0.000 |
| MyVoteSys <$\alpha$=0.66, $\beta$=0.33, $\gamma$=0.80, MinVariance> | 1.470 | 0.706 | 0.000 |
| MyVoteSys <$\alpha$=0.66, $\beta$=____, $\gamma$=____, MinVariance> | 1.473 | 0.694 | 0.000 |
| MyVoteSys <$\alpha$=0.66, $\beta$=0.33, $\gamma$=____, MinVariance> | 1.481 | 0.700 | 0.000 |
| MyVoteSys <$\alpha$=0.66, $\beta$=0.33, $\gamma$=0.80, MaxEntropy> | 1.482 | 0.697 | 0.000 |
| MyVoteSys <$\alpha$=0.66, $\beta$=0.33, $\gamma$=0.80, Last> | 1.482 | 0.697 | 0.000 |
| MaxStandardDeviationWeightedSumWins<$\alpha$=0.5> | 1.485 | 0.685 | 0.000 |
| MyVoteSys <$\alpha$=0.66, $\beta$=0.33, $\gamma$=____, Last> | 1.501 | 0.684 | 0.000 |
| MyVoteSys <$\alpha$=0.66, $\beta$=0.33, $\gamma$=____, MaxEntropy> | 1.501 | 0.684 | 0.000 |
| MyVoteSys <$\alpha$=0.50, $\beta$=____, $\gamma$=0.80, MaxEntropy> | 1.537 | 0.681 | 0.000 |
| MyVoteSys <$\alpha$=0.80, $\beta$=____, $\gamma$=____, MinEntropy> | 1.537 | 0.681 | 0.000 |
| MyVoteSys <$\alpha$=0.50, $\beta$=____, $\gamma$=0.80, Last> | 1.537 | 0.681 | 0.000 |
| MyVoteSys <$\alpha$=0.66, $\beta$=____, $\gamma$=0.80, Last> | 1.537 | 0.681 | 0.000 |
| MyVoteSys <$\alpha$=0.80, $\beta$=____, $\gamma$=____, First> | 1.537 | 0.681 | 0.000 |
| MyVoteSys <$\alpha$=0.66, $\beta$=____, $\gamma$=0.80, MaxEntropy> | 1.537 | 0.681 | 0.000 |
| MaxVarianceProportionalWeightedSumWins<$\alpha$=0> | 1.542 | 0.656 | 0.000 |
| MaxVarianceProportionalWeightedSumWins<$\alpha$=0.5> | 1.552 | 0.652 | 0.000 |
| MaxStandardDeviationProportionalWeightedSumWins<$\alpha$=0> | 1.570 | 0.649 | 0.000 |
| MyVoteSys <$\alpha$=0.80, $\beta$=____, $\gamma$=____, MinVariance> | 1.584 | 0.668 | 0.000 |
| MaxStandardDeviationProportionalWeightedSumWin<$\alpha$=0.5> | 1.584 | 0.645 | 0.000 |
| FirstPastThePost (but without tactical voting) | 1.966 | 0.581 | 0.001 |
| MyVoteSys <$\alpha$=0.80, $\beta$=____, $\gamma$=____, MaxVariance> | 2.862 | 0.330 | 0.018 |
| MyVoteSys <$\alpha$=0.66, $\beta$=____, $\gamma$=____, Last> | 2.862 | 0.330 | 0.018 |
| MyVoteSys <$\alpha$=0.80, $\beta$=____, $\gamma$=____, MaxStDev> | 2.862 | 0.330 | 0.018 |
| MyVoteSys <$\alpha$=0.50, $\beta$=____, $\gamma$=____, MaxEntropy> | 2.862 | 0.330 | 0.018 |
| MyVoteSys <$\alpha$=0.66, $\beta$=____, $\gamma$=____, MaxStDev> | 2.862 | 0.330 | 0.018 |
| MyVoteSys <$\alpha$=0.80, $\beta$=____, $\gamma$=____, MaxEntropy> | 2.862 | 0.330 | 0.018 |
| MyVoteSys <$\alpha$=0.50, $\beta$=____, $\gamma$=____, MaxVariance> | 2.862 | 0.330 | 0.018 |
| MyVoteSys <$\alpha$=0.50, $\beta$=____, $\gamma$=____, Last> | 2.862 | 0.330 | 0.018 |
| MyVoteSys <$\alpha$=0.66, $\beta$=____, $\gamma$=____, MaxEntropy> | 2.862 | 0.330 | 0.018 |
| MyVoteSys <$\alpha$=0.80, $\beta$=____, $\gamma$=____, Last> | 2.862 | 0.330 | 0.018 |
| MyVoteSys <$\alpha$=0.50, $\beta$=____, $\gamma$=____, MaxStDev> | 2.862 | 0.330 | 0.018 |
| MyVoteSys <$\alpha$=0.80, $\beta$=0.33, $\gamma$=____, First> | 3.142 | 0.585 | 0.000 |
| MyVoteSys <$\alpha$=0.80, $\beta$=0.33, $\gamma$=____, MinEntropy> | 3.142 | 0.585 | 0.000 |
| MyVoteSys <$\alpha$=0.80, $\beta$=0.33, $\gamma$=____, First> | 3.152 | 0.577 | 0.000 |
| MyVoteSys <$\alpha$=0.80, $\beta$=0.33, $\gamma$=____, MaxStDev> | 3.152 | 0.577 | 0.000 |
| MyVoteSys <$\alpha$=0.80, $\beta$=0.33, $\gamma$=____, MinVariance> | 3.153 | 0.578 | 0.000 |
| MyVoteSys <$\alpha$=0.80, $\beta$=0.33, $\gamma$=____, MaxEntropy> | 3.161 | 0.572 | 0.000 |
| MyVoteSys <$\alpha$=0.80, $\beta$=0.33, $\gamma$=____, Last> | 3.161 | 0.572 | 0.000 |
| bestVoter (Dictatorship) | 4.148 | 0.285 | 0.086 |
| MinEntropyWeightedSumWins2<$\alpha$=0> | 5.361 | 0.169 | 0.128 |
| MinEntropyWeightedSumWins<$\alpha$=0> | 5.361 | 0.169 | 0.128 |
| MinEntropyProportionalWeightedSumWins2<$\alpha$=0> | 5.361 | 0.169 | 0.128 |
| MinEntropyProportionalWeightedSumWins<$\alpha$=0> | 5.361 | 0.169 | 0.128 |
| MinEntropyProportionalWeightedSumWins<$\alpha$=0.5> | 5.365 | 0.169 | 0.128 |
| MinEntropyWeightedSumWins<$\alpha$=0.5> | 5.365 | 0.169 | 0.128 |
| MinEntropyWeightedSumWins2<$\alpha$=0.5> | 5.365 | 0.169 | 0.128 |
| MinEntropyProportionalWeightedSumWins2<$\alpha$=0.5> | 5.365 | 0.169 | 0.128 |

| Metrics | val_MeanSquaredErr |
|---|---|
| Algorithms | |
| bestVoter (Dictatorship) | 11196.712891 |
| crowd-Median | 17084.095703 |
| crowd-Mean | 17188.355469 |

================================ END OF SIMULATION 8 ================================

================================ START OF SIMULATION 9 ================================

numCandiates : 20
numVoters : 1000
numElections : 1000
columnBlindness : [4, 6]
epochs : 133
trainableLayerCount : 1
crowdBuildMethod : {'name': 'standardDistribution', 'mean': 10500, 'standardDeviation': 1833}
dataSetName : mySynthetic
predictedFeature : y
========= SIMULATION RESULTS =======

| Metrics | meanWinnerRank | rateTrueWinners | rateWinner<NULL |
|---|---|---|---|
| Algorithms | | | |
| crowd-Mean | 1.065 | 0.943 | 0.000 |
| crowd-Median | 1.065 | 0.943 | 0.000 |
| MyVoteSys <$\alpha$=0.50, $\beta$=____, $\gamma$=0.66, MaxVariance> | 1.090 | 0.918 | 0.000 |
| MyVoteSys <$\alpha$=0.50, $\beta$=0.33, $\gamma$=0.80, First> | 1.090 | 0.918 | 0.000 |
| MyVoteSys <$\alpha$=0.50, $\beta$=0.33, $\gamma$=____, MinEntropy> | 1.090 | 0.918 | 0.000 |
| MyVoteSys <$\alpha$=0.50, $\beta$=0.33, $\gamma$=____, First> | 1.090 | 0.918 | 0.000 |
| MyVoteSys <$\alpha$=0.50, $\beta$=____, $\gamma$=0.80, First> | 1.090 | 0.918 | 0.000 |
| MyVoteSys <$\alpha$=0.50, $\beta$=____, $\gamma$=____, First> | 1.090 | 0.918 | 0.000 |
| MyVoteSys <$\alpha$=0.50, $\beta$=____, $\gamma$=0.66, MinEntropy> | 1.090 | 0.918 | 0.000 |
| MyVoteSys <$\alpha$=0.50, $\beta$=____, $\gamma$=0.80, MinEntropy> | 1.090 | 0.918 | 0.000 |
| MyVoteSys <$\alpha$=0.50, $\beta$=0.33, $\gamma$=0.80, MinEntropy> | 1.090 | 0.918 | 0.000 |
| MyVoteSys <$\alpha$=0.50, $\beta$=0.33, $\gamma$=0.66, MaxVariance> | 1.090 | 0.918 | 0.000 |
| MyVoteSys <$\alpha$=0.50, $\beta$=____, $\gamma$=0.80, MaxVariance> | 1.090 | 0.918 | 0.000 |
| MyVoteSys <$\alpha$=0.50, $\beta$=0.33, $\gamma$=0.66, MinEntropy> | 1.090 | 0.918 | 0.000 |
| MyVoteSys <$\alpha$=0.50, $\beta$=0.33, $\gamma$=0.80, MaxVariance> | 1.090 | 0.918 | 0.000 |
| MyVoteSys <$\alpha$=0.50, $\beta$=0.33, $\gamma$=0.66, First> | 1.090 | 0.918 | 0.000 |
| MyVoteSys <$\alpha$=0.50, $\beta$=0.33, $\gamma$=____, MaxVariance> | 1.090 | 0.918 | 0.000 |
| MyVoteSys <$\alpha$=0.50, $\beta$=____, $\gamma$=____, MinEntropy> | 1.090 | 0.918 | 0.000 |
| MyVoteSys <$\alpha$=0.50, $\beta$=____, $\gamma$=0.66, First> | 1.090 | 0.918 | 0.000 |
| InstantRunoffVoting | 1.095 | 0.925 | 0.000 |
| PreferentialBlockVoting | 1.096 | 0.925 | 0.000 |
| SingleTransferableVote | 1.097 | 0.924 | 0.000 |
| MyVoteSys <$\alpha$=0.50, $\beta$=0.33, $\gamma$=0.66, MinVariance> | 1.124 | 0.888 | 0.000 |
| MyVoteSys <$\alpha$=0.66, $\beta$=0.33, $\gamma$=0.80, MaxVariance> | 1.124 | 0.888 | 0.000 |
| MyVoteSys <$\alpha$=0.50, $\beta$=____, $\gamma$=0.66, MaxEntropy> | 1.124 | 0.888 | 0.000 |

| | | | | | | |
|---|---|---|---|---|---|---|
| MyVoteSys | ⟨α=0.66, | β=0.33, | γ=0.80, | MinEntropy⟩ | 1.124 | 0.888 | 0.000 |
| MyVoteSys | ⟨α=0.50, | β=0.33, | γ=0.66, | MaxEntropy⟩ | 1.124 | 0.888 | 0.000 |
| MyVoteSys | ⟨α=0.66, | β=0.33, | γ=0.80, | First⟩ | 1.124 | 0.888 | 0.000 |
| MyVoteSys | ⟨α=0.50, | β=0.33, | γ=0.66, | Last⟩ | 1.124 | 0.888 | 0.000 |
| MyVoteSys | ⟨α=0.66, | β=0.33, | γ=____, | MinEntropy⟩ | 1.124 | 0.888 | 0.000 |
| MyVoteSys | ⟨α=0.50, | β=____, | γ=0.66, | MinVariance⟩ | 1.124 | 0.888 | 0.000 |
| MyVoteSys | ⟨α=0.66, | β=____, | γ=0.80, | First⟩ | 1.124 | 0.888 | 0.000 |
| MyVoteSys | ⟨α=0.50, | β=____, | γ=0.66, | Last⟩ | 1.124 | 0.888 | 0.000 |
| MyVoteSys | ⟨α=0.66, | β=0.33, | γ=____, | First⟩ | 1.124 | 0.888 | 0.000 |
| MyVoteSys | ⟨α=0.66, | β=____, | γ=____, | MinEntropy⟩ | 1.124 | 0.888 | 0.000 |
| MyVoteSys | ⟨α=0.66, | β=____, | γ=0.80, | MaxVariance⟩ | 1.124 | 0.888 | 0.000 |
| MyVoteSys | ⟨α=0.66, | β=____, | γ=0.80, | MinEntropy⟩ | 1.124 | 0.888 | 0.000 |
| MyVoteSys | ⟨α=0.66, | β=____, | γ=____, | First⟩ | 1.124 | 0.888 | 0.000 |
| MyVoteSys | ⟨α=0.66, | β=0.33, | γ=____, | MaxVariance⟩ | 1.126 | 0.887 | 0.000 |
| FirstPastThePost (but without tactical voting) | | | | | 1.195 | 0.856 | 0.000 |
| MyVoteSys | ⟨α=0.66, | β=0.33, | γ=0.80, | MinVariance⟩ | 1.222 | 0.827 | 0.000 |
| MyVoteSys | ⟨α=0.66, | β=0.33, | γ=0.80, | Last⟩ | 1.222 | 0.827 | 0.000 |
| MyVoteSys | ⟨α=0.66, | β=0.33, | γ=0.80, | MaxEntropy⟩ | 1.222 | 0.827 | 0.000 |
| MyVoteSys | ⟨α=0.50, | β=0.33, | γ=0.80, | MaxEntropy⟩ | 1.222 | 0.827 | 0.000 |
| MyVoteSys | ⟨α=0.50, | β=0.33, | γ=0.80, | MinVariance⟩ | 1.222 | 0.827 | 0.000 |
| MyVoteSys | ⟨α=0.50, | β=0.33, | γ=0.80, | Last⟩ | 1.222 | 0.827 | 0.000 |
| MyVoteSys | ⟨α=0.66, | β=____, | γ=0.80, | Last⟩ | 1.226 | 0.825 | 0.000 |
| MyVoteSys | ⟨α=0.80, | β=____, | γ=____, | First⟩ | 1.226 | 0.825 | 0.000 |
| MyVoteSys | ⟨α=0.80, | β=____, | γ=____, | MinEntropy⟩ | 1.226 | 0.825 | 0.000 |
| MyVoteSys | ⟨α=0.50, | β=____, | γ=0.80, | MaxEntropy⟩ | 1.226 | 0.825 | 0.000 |
| MyVoteSys | ⟨α=0.66, | β=____, | γ=0.80, | MaxEntropy⟩ | 1.226 | 0.825 | 0.000 |
| MyVoteSys | ⟨α=0.66, | β=____, | γ=0.80, | MinVariance⟩ | 1.226 | 0.825 | 0.000 |
| MyVoteSys | ⟨α=0.50, | β=____, | γ=0.80, | Last⟩ | 1.226 | 0.825 | 0.000 |
| MyVoteSys | ⟨α=0.50, | β=____, | γ=0.80, | MinVariance⟩ | 1.226 | 0.825 | 0.000 |
| MyVoteSys | ⟨α=0.66, | β=0.33, | γ=____, | MinVariance⟩ | 1.327 | 0.757 | 0.000 |
| MyVoteSys | ⟨α=0.50, | β=0.33, | γ=____, | MinVariance⟩ | 1.327 | 0.757 | 0.000 |
| MyVoteSys | ⟨α=0.50, | β=0.33, | γ=____, | Last⟩ | 1.356 | 0.740 | 0.000 |
| MyVoteSys | ⟨α=0.66, | β=0.33, | γ=____, | Last⟩ | 1.356 | 0.740 | 0.000 |
| MyVoteSys | ⟨α=0.66, | β=0.33, | γ=____, | MaxEntropy⟩ | 1.356 | 0.740 | 0.000 |
| MyVoteSys | ⟨α=0.50, | β=0.33, | γ=____, | MaxEntropy⟩ | 1.356 | 0.740 | 0.000 |
| MyVoteSys | ⟨α=0.80, | β=0.33, | γ=____, | First⟩ | 1.372 | 0.811 | 0.000 |
| MyVoteSys | ⟨α=0.80, | β=0.33, | γ=____, | MinEntropy⟩ | 1.372 | 0.811 | 0.000 |
| MyVoteSys | ⟨α=0.80, | β=0.33, | γ=____, | MaxVariance⟩ | 1.379 | 0.810 | 0.000 |
| MyVoteSys | ⟨α=0.66, | β=____, | γ=____, | MinVariance⟩ | 1.404 | 0.730 | 0.000 |
| MyVoteSys | ⟨α=0.50, | β=____, | γ=____, | MinVariance⟩ | 1.404 | 0.730 | 0.000 |
| MyVoteSys | ⟨α=0.80, | β=____, | γ=____, | MinVariance⟩ | 1.404 | 0.730 | 0.000 |
| MyVoteSys | ⟨α=0.80, | β=0.33, | γ=____, | MinVariance⟩ | 1.477 | 0.741 | 0.000 |
| MyVoteSys | ⟨α=0.80, | β=0.33, | γ=____, | Last⟩ | 1.506 | 0.724 | 0.000 |
| MyVoteSys | ⟨α=0.80, | β=0.33, | γ=____, | MaxEntropy⟩ | 1.506 | 0.724 | 0.000 |
| MyVoteSys | ⟨α=0.50, | β=____, | γ=____, | MaxVariance⟩ | 1.748 | 0.607 | 0.003 |
| MyVoteSys | ⟨α=0.66, | β=____, | γ=____, | MaxVariance⟩ | 1.756 | 0.604 | 0.003 |
| MyVoteSys | ⟨α=0.80, | β=____, | γ=____, | Last⟩ | 1.760 | 0.603 | 0.003 |
| MyVoteSys | ⟨α=0.50, | β=____, | γ=____, | Last⟩ | 1.760 | 0.603 | 0.003 |
| MyVoteSys | ⟨α=0.50, | β=____, | γ=____, | MaxEntropy⟩ | 1.760 | 0.603 | 0.003 |
| MyVoteSys | ⟨α=0.66, | β=____, | γ=____, | MaxEntropy⟩ | 1.760 | 0.603 | 0.003 |
| MyVoteSys | ⟨α=0.80, | β=____, | γ=____, | MaxEntropy⟩ | 1.760 | 0.603 | 0.003 |
| MyVoteSys | ⟨α=0.80, | β=____, | γ=____, | MaxVariance⟩ | 1.760 | 0.603 | 0.003 |
| MyVoteSys | ⟨α=0.66, | β=____, | γ=____, | Last⟩ | 1.760 | 0.603 | 0.003 |
| bestVoter (Dictatorship) | | | | | 2.592 | 0.425 | 0.002 |

| Metrics | val_MeanSquaredErr |
|---|---|
| Algorithms | |
| crowd-Mean | 19522.767578 |
| crowd-Median | 19218.832031 |
| bestVoter (Dictatorship) | 6214.651855 |

================================ END OF SIMULATION 9 ================================

## IV. Remarks

• The complexity of the algorithm is obviously $O(n^2)$, after counting the votes of course ($n$ is the number of candidates; note that $f_{1(X,i)} = f_{1(X,i-1)} + x_i$).

• When enough stages, the voter has no reason to resort to tactical voting.

• An abstraction of $\gamma$ can have, to name a few, the following concretions: $\gamma = 0.8$, $\gamma_{0.5} = 0.9$, $\gamma_2 = 0.95$ (where $\gamma = 0.8$ makes stages invalid if any candidate exceeds the score 80%, $\gamma_{0.5} = 0.9$ makes stages invalid if at least 50% of candidates exceed the score 0.9%, and $\gamma_2 = 0.95$ makes stages invalid if at least 2 candidates exceed the score of 95%).

• It is very important that voters, are making educated guesses. If voters submit incomplete ballots, with just one preference, because of ignorance, then this is very unfortunate, because we want as much heterogeneous and sincere contribution as possible. It is not wrong to not know which candidate is better. As long a stamp is the result of an independent, educated guess, it's a valuable contribution.

• Even if conditions are met for supporting the wisdom of the crowds, we need to consider that if the best voter is a very far outlier (for example when the best voter is Einstein and all the other voters are not much better than chimpanzees), or if nobody is interested in democracy, then maybe dictatorship is better.

• Based on the simulation results, a version tends to work better if they have $\alpha \in [0.50, 0.66]$, $\beta = 0.33$, high or no $\gamma$, and minimum entropy or maximum variance as win-stage-selector (over the pool of consecutive valid stages). Also, versions where the gap between $\alpha$ and $\gamma$ is small, and rely on maximum entropy or minimum variance, tend to be the worst.

• In simulation 8 and 9, voter blindness is a number from an interval, and for the one with interval [7,9].

• One interesting thing, that we are not able to explain, is that, even if the best voter is much better at regression (mean squared error used) than Crowd-Mean and Crowd-Median, it is still vastly outperformed in voting. To us, this is very unexpected, and we couldn't even find a bug in the software.

• Simulation results may differ depending on how the qualities of voters are distributed. We only care about a distribution that is close to a normal distribution, because it's natural.

• It should come to no surprise to see that crowd-Mean and crowd-Median are by far the best, since this voting system provides full freedom of expression (because each candidate is assessed with real numbers, numbers that hold more information than ranks). However we generally don't recommend this kind of voting system, for people to vote, because people are likely to lie and resort to tactical voting. Obviously, a ranked voting system, forces the voter, to be sincere, in assessing each candidate.

• In the simulations, we sorted the columns (candidates) of "score of candidates table", in decreasing order of score, with higher stage as priority. We did not put much thought into this, but it is important when two candidates have equal best score on the same stage.

Here are some of the major characteristics of the presented voting system(s):

- Voters have freedom of expression on their ballots, without loss or dismissal of provided information.
- The NULL candidate $\varnothing$, is introduced, which significantly improves the capacity to better understand what the voters want.
- A hyper-parameter ($\beta$) can be used to minimise the number of discontented voters.
- A candidate might win if he is nobody's first favorite but is everyone's second favorite.
- As expected, for a ranked voting system, it is harder, for a voter to be afraid of wasting his vote/stamps, meaning less pressure on voters to become insincere and resort to tactical voting.
- We can be creative by deriving the algorithm.

## V. Results

We presented a novel ranked voting system that is highly configurable (has parameters). Simulation results show that, with the right parameters, the novel voting system, is easily the better option, if the conditions supporting the wisdom of the crowds, are met. In other words, simulation results show that the novel voting system is better than Instant-Runoff Voting, Preferential Block Voting, Single Transferable Vote, and First Past The Post. By also comparing with the best voter, the simulations also demonstrate the wisdom of the crowds, which suggests that democracy (distributed system) is a better choice, than dictatorship (centralized system), in an environment that facilitates the wisdom of the crowds.

APPENDIX

*A. Random Simulations*

In this chapter, we will present results after random, unrealistic, simulations with two versions of the algorithm. The source code, used to make these random simulations, is available on GitHub [1] as first commit.

The ballots are generated randomly, so the examples shown are cherry-picked by the application, to show how it behaves in different situations. Candidates are represented by numbers (first row of the table). Note that stage numbers begin from 0.

One version of the algorithm is the basic one, which only relies on $\alpha = 0.5$. The other version relies on $\alpha = 0.5$, $\beta = 0.3333$, $\gamma = 0.6666$ as explained at section III-D.

| Simulation 1 | Score of Candidates Table: | | | | | | | |
|---|---|---|---|---|---|---|---|---|
| | | 0 | 1 | 2 | 3 | 4 | 5 | 6 | 7 |
| | 0 | 18.75% | 12.50% | 18.75% | 6.25% | 6.25% | 18.75% | 18.75% | 0.00% |
| | 1 | 18.75% | 25.00% | 31.25% | 31.25% | 6.25% | 31.25% | 18.75% | 37.50% |
| | 2 | 18.75% | 37.50% | 37.50% | 31.25% | 43.75% | 31.25% | 37.50% | 62.50% |
| | 3 | 18.75% | 43.75% | 62.50% | 56.25% | 50.00% | 43.75% | 50.00% | 75.00% |
| | 4 | 25.00% | 50.00% | 62.50% | 87.50% | 81.25% | 50.00% | 62.50% | 81.25% |
| | 5 | 37.50% | 68.75% | 68.75% | 100.00% | 93.75% | 68.75% | 81.25% | 81.25% |
| | 6 | 62.50% | 81.25% | 87.50% | 100.00% | 93.75% | 87.50% | 87.50% | 100.00% |
| | 7 | 100.00% | 100.00% | 100.00% | 100.00% | 100.00% | 100.00% | 100.00% | 100.00% |

| Basic Algorithm | BetaGamma Algorithm | | |
|---|---|---|---|
| algorithm: basic | algorithm: BetaGamma | bestScoreStage: 3 | firstStageByAlpha: 2 |
| winner: 7 | NULLCandidate: 0 | lastStageByBeta: 4 | bestScoreByAlpha: 62.5 |
| stage: 2 | winner: 7 | bestScoreByBeta: 87.5 | alpha: 0.5 |
| score: 62.5 | bestCandidate: 7 | lastStageByGamma: 3 | beta: 0.3333 |
| alpha: 0.5 | bestScore: 75.0 | bestScoreByGamma: 75.0 | gamma: 0.6666 |

The variable lastStageByBeta is referring to the last stage at which a winner can be selected according to $\beta$. The variable bestScoreStage is referring to the last stage at which a winner can be selected according to both $\gamma$ and $\beta$. The other variables are self explanatory in the same manner.

We remind that we do not want to select a winner on a stage where $\varnothing$ (NULLCandidate) passes the threshold dictated by $\beta$, since we use $\beta$ to minimize the number of unsatisfied voters.

NULLCandidate ($\varnothing$) is on the column labeled with 0.

| Simulation 2 | Score of Candidates Table: | | | | | | | |
|---|---|---|---|---|---|---|---|---|
| | | 0 | 1 | 2 | 3 | 4 | 5 | 6 | 7 |
| | 0 | 6.25% | 12.50% | 12.50% | 12.50% | 6.25% | 12.50% | 31.25% | 6.25% |
| | 1 | 12.50% | 25.00% | 31.25% | 18.75% | 6.25% | 31.25% | 56.25% | 18.75% |
| | 2 | 12.50% | 37.50% | 50.00% | 25.00% | 12.50% | 43.75% | 68.75% | 50.00% |
| | 3 | 18.75% | 56.25% | 62.50% | 37.50% | 37.50% | 62.50% | 75.00% | 50.00% |
| | 4 | 37.50% | 75.00% | 62.50% | 50.00% | 56.25% | 68.75% | 93.75% | 56.25% |
| | 5 | 62.50% | 81.25% | 93.75% | 68.75% | 56.25% | 81.25% | 93.75% | 62.50% |
| | 6 | 81.25% | 93.75% | 100.00% | 93.75% | 75.00% | 87.50% | 93.75% | 75.00% |
| | 7 | 100.00% | 100.00% | 100.00% | 100.00% | 100.00% | 100.00% | 100.00% | 100.00% |

| Basic Algorithm | BetaGamma Algorithm | | |
|---|---|---|---|
| algorithm: basic | algorithm: BetaGamma | bestScoreStage: 2 | firstStageByAlpha: 1 |
| winner: 6 | NULLCandidate: 0 | lastStageByBeta: 3 | bestScoreByAlpha: 56.25 |
| stage: 1 | winner: 6 | bestScoreByBeta: 75.0 | alpha: 0.5 |
| score: 56.25 | bestCandidate: 6 | lastStageByGamma: 2 | beta: 0.3333 |
| alpha: 0.5 | bestScore: 68.75 | bestScoreByGamma: 68.75 | gamma: 0.6666 |

## Simulation 3

Score of Candidates Table:

|   | 0 | 1 | 2 | 3 | 4 | 5 | 6 | 7 |
|---|---|---|---|---|---|---|---|---|
| 0 | 6.25% | 6.25% | 6.25% | 6.25% | 18.75% | 18.75% | 25.00% | 12.50% |
| 1 | 18.75% | 37.50% | 12.50% | 18.75% | 18.75% | 37.50% | 31.25% | 25.00% |
| 2 | 31.25% | 43.75% | 43.75% | 37.50% | 18.75% | 25.00% | 56.25% | 50.00% |
| 3 | 43.75% | 56.25% | 50.00% | 25.00% | 50.00% | 68.75% | 56.25% | 50.00% |
| 4 | 43.75% | 62.50% | 62.50% | 43.75% | 68.75% | 81.25% | 62.50% | 75.00% |
| 5 | 56.25% | 68.75% | 81.25% | 56.25% | 81.25% | 93.75% | 81.25% | 81.25% |
| 6 | 100.00% | 81.25% | 93.75% | 62.50% | 87.50% | 100.00% | 87.50% | 87.50% |
| 7 | 100.00% | 100.00% | 100.00% | 100.00% | 100.00% | 100.00% | 100.00% | 100.00% |

| Basic Algorithm | BetaGamma Algorithm |
|---|---|

algorithm: basic
winner: 5
stage: 2
score: 56.25
alpha: 0.5

algorithm: BetaGamma    bestScoreStage: 2    firstStageByAlpha: 2
NULLCandidate: 0    lastStageByBeta: 2    bestScoreByAlpha: 56.25
winner: 5    bestScoreByBeta: 56.25    alpha: 0.5
bestCandidate: 5    lastStageByGamma: 3    beta: 0.3333
bestScore: 56.25    bestScoreByGamma: 68.75    gamma: 0.6666

## Simulation 4

Score of Candidates Table:

|   | 0 | 1 | 2 | 3 | 4 | 5 | 6 | 7 |
|---|---|---|---|---|---|---|---|---|
| 0 | 6.25% | 6.25% | 18.75% | 25.00% | 6.25% | 12.50% | 12.50% | 12.50% |
| 1 | 37.50% | 18.75% | 43.75% | 31.25% | 18.75% | 12.50% | 12.50% | 25.00% |
| 2 | 43.75% | 25.00% | 50.00% | 43.75% | 50.00% | 31.25% | 25.00% | 31.25% |
| 3 | 50.00% | 43.75% | 68.75% | 62.50% | 56.25% | 43.75% | 37.50% | 37.50% |
| 4 | 50.00% | 56.25% | 68.75% | 81.25% | 75.00% | 62.50% | 56.25% | 50.00% |
| 5 | 62.50% | 62.50% | 87.50% | 87.50% | 87.50% | 75.00% | 68.75% | 68.75% |
| 6 | 75.00% | 75.00% | 93.75% | 93.75% | 93.75% | 93.75% | 87.50% | 87.50% |
| 7 | 100.00% | 100.00% | 100.00% | 100.00% | 100.00% | 100.00% | 100.00% | 100.00% |

| Basic Algorithm | BetaGamma Algorithm |
|---|---|

algorithm: basic
winner: 2
stage: 3
score: 68.75
alpha: 0.5

algorithm: BetaGamma    bestScoreStage: 0    firstStageByAlpha: 3
NULLCandidate: 0    lastStageByBeta: 0    bestScoreByAlpha: 68.75
winner: 0    bestScoreByBeta: 25.0    alpha: 0.5
bestCandidate: 3    lastStageByGamma: 3    beta: 0.3333
bestScore: 25.0    bestScoreByGamma: 68.75    gamma: 0.6666

## Simulation 5

Score of Candidates Table:

|   | 0 | 1 | 2 | 3 | 4 | 5 | 6 | 7 |
|---|---|---|---|---|---|---|---|---|
| 0 | 18.75% | 12.50% | 18.75% | 0.00% | 18.75% | 6.25% | 12.50% | 12.50% |
| 1 | 18.75% | 37.50% | 31.25% | 6.25% | 25.00% | 25.00% | 25.00% | 31.25% |
| 2 | 25.00% | 50.00% | 37.50% | 31.25% | 31.25% | 31.25% | 56.25% | 37.50% |
| 3 | 31.25% | 75.00% | 50.00% | 37.50% | 56.25% | 43.75% | 62.50% | 43.75% |
| 4 | 31.25% | 87.50% | 62.50% | 50.00% | 68.75% | 62.50% | 87.50% | 50.00% |
| 5 | 62.50% | 87.50% | 81.25% | 68.75% | 87.50% | 75.00% | 87.50% | 50.00% |
| 6 | 87.50% | 87.50% | 87.50% | 75.00% | 100.00% | 87.50% | 93.75% | 81.25% |
| 7 | 100.00% | 100.00% | 100.00% | 100.00% | 100.00% | 100.00% | 100.00% | 100.00% |

| Basic Algorithm | BetaGamma Algorithm |
|---|---|

algorithm: basic
winner: 6
stage: 2
score: 56.25
alpha: 0.5

algorithm: BetaGamma    bestScoreStage: 3    firstStageByAlpha: 2
NULLCandidate: 0    lastStageByBeta: 4    bestScoreByAlpha: 56.25
winner: 1    bestScoreByBeta: 87.5    alpha: 0.5
bestCandidate: 1    lastStageByGamma: 3    beta: 0.3333
bestScore: 75.0    bestScoreByGamma: 75.0    gamma: 0.6666

## B. Failed attempt of a better version

One version of the algorithm, is calculating a weighted sum of points that a candidate receives each time he wins on a stage (the sum is the score). The weight is actually the entropy of the stage. We also tried multiplying the weight with the stage number. We tried variance and standard, distribution in staid of entropy. We also tried a version, subtracting entropy from the maximum entropy, and $\alpha = 0.5$. Unfortunately, they don't perform well at all, maybe something is missing. We did not test these variations, at every simulation.

## C. Analogous descriptions of the basic algorithm

For simplicity, we will not mention $\varnothing$ (the null candidate/vote), but it should be noted that it is an important piece of the algorithm.

- Each day, farmers distribute the same amount of $n$ seeds, in $k$ barrels. We don't know how many seeds each barrel will be supplied with each day, but we want to stop, on the day when there is at least 1 barrel, that has more than $\alpha \cdot n$ seeds. $\alpha$ being arbitrary, $\alpha \in [0,1]$ but preferably $\alpha = 0.5$ (to better represent the voters).
  They are $n$ farmers so this means 1 seed per farmer. A farmer can only supply each barrel once.
  In our voting problem, farmers are voters, the barrels are candidates, the days are stages, the seeds are votes (each seed is a stamp).

- We have $k$ horses competing with each other, in a race. The distance they need to travel, in order to win, is $\alpha \cdot n$. But the horses cannot control them selves, and each of their steps have the same length of 1 unit. The horses are remotely controlled by n supporters. A supporter can only control a particular horse once, by advancing the horse with 1 step. Each second, all the supporters are, simultaneously, controlling the horses. If at any second, $p$ supporters are controlling the same horse, then that horse will make $p$ steps. At the finish line, there is a post. The first horse past the post wins.
  In our voting problem, horses are candidates, the seconds are stages, the remotely controlled steps are stamps, the supporters are voters.

## D. Ballot design

We could use rectangles, to associate a candidate, in each of our ballots, as follows:

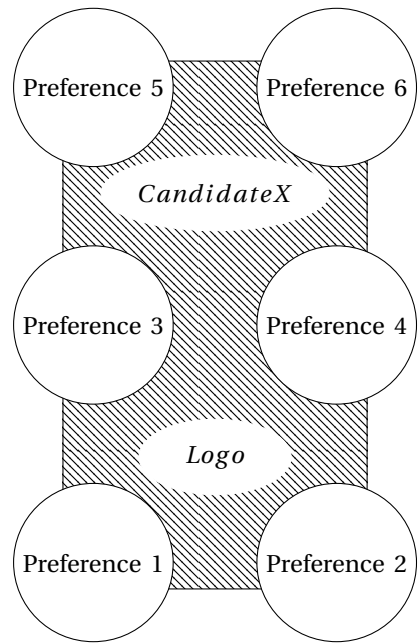

But here is a more practical ballot design which is already used in ranked voting systems:

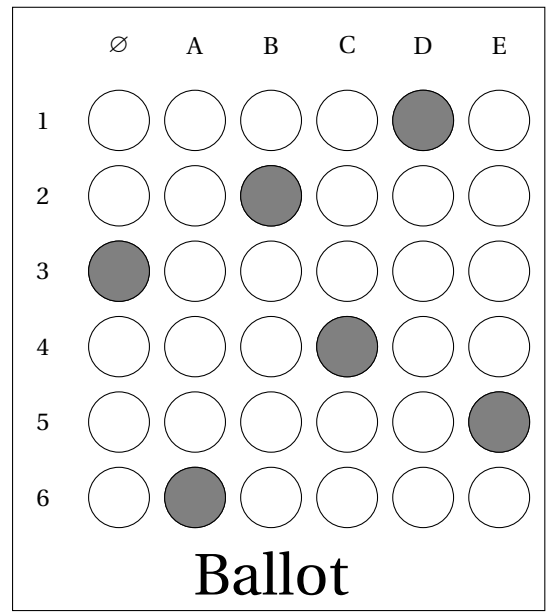

The above ballot can be interpreted as:
D > B > $\varnothing$ > C > E > A .

## E. Dealing with incomplete ballots

First of all, each missing stamp is effectively a vote for the "I don't know" candidate.

If a voter submitted a ballot Q, that is not complete, then he does not give any advantage to the non-stamped candidates (from his ballot Q). But, those that do submit a complete ballot, as a result, give an advantage to all the candidates, whether they like it or not. It is not fair to only give advantage to some of the candidates, when others offer advantage to all of the candidates.

However, there is a fair way to deal with incomplete ballots, without removing any incomplete ballots (at any stage). The solution is to default each missing preference to a value of $\frac{1}{k}$ where $k$ is the number of non-stamped candidates from the respective ballot. This means that, the vote count $x_i$, can be any positive number, not just a natural number.

So, for example, if a voter submits a ballot with no stamp on it, then it should mean that all the options are equal to him, thus, every candidate receives the same amount of credit at every preference, as if the voter submitted a complete ballot, but each of his stamps are split evenly in all the stamping locations of the same preference.

*F. Reducing the number of stages*

- If every voter is submitting an incomplete ballot, containing just 1 candidate, and $\alpha=0$, then the algorithm is reduced to First Past The Post voting system, which we already know it is often unfair, because the winner may not represent the majority of voters and because in time, it usually and inevitably, leads to the same 2 major (party) candidates (other known disadvantages do exist).

This suggests that if the ballots support $k$ preferences, in time, usually and inevitably, it will lead to the same $k+1$ major (party) candidates.

- If we reduce the number of stages, it is possible that no candidate will pass the threshold, in which case we have no choice but to ignore $\alpha$, at the last stage, and pick the candidate with the highest score (which introduces some unfairness, related to that of First Past The Post algorithm).

- Obviously, the last possible stage is not necessary, because it should always lead to scores of 100%, which means voters don't need to submit complete ballots.

*G. Minimum number of necessary stages (or preferences supported by ballots)*

If all possible stages exist, and if they are enough voters, a score of above 50% is guaranteed to happen before the last possible stage. As proof, we can consider the most unfortunate example, as represented in the next table:

| Vote Counts | | | | |
|---|---|---|---|---|
| | A | B | C | D | $\varnothing$ |
| Preference 1 | 20 | 20 | 20 | 20 | 20 |
| Preference 2 | 20 | 20 | 20 | 20 | 20 |
| Preference 3 | 20 | 20 | 20 | 20 | 20 |
| Preference 4 | 20 | 20 | 20 | 20 | 20 |
| Preference 5 | 20 | 20 | 20 | 20 | 20 |

Lets calculate the minimum number of stages needed to make sure there is at least one candidate that passes the threshold $\alpha$:

$$\frac{n}{k}x > \alpha n \Leftrightarrow x > \alpha k$$

$n$ = the number of voters
$k$ = the number of candidates
$x$ = the (minimum) number of stages
$\alpha$ = the threshold $(0 < \alpha < 1)$

In our example, based on the previous table, we observe that $n = 100$ and $k = 5$. If we select an arbitrary $\alpha = 0.5$, and use the previous equation, we obtain:

$$x > 0.5 \cdot 5 \Leftrightarrow x > 2.5$$

$$x > 2.5 \text{ and } x \in \mathbb{N} \Rightarrow x = 3$$

However, we may need to add one more stage, every time, they are top candidates that ended up having equal score, or look in the past stage, which of the candidates had the most advantage. In cases where the number of voters is big enough, it is reasonable to risk, and settle with the minimum number of stages calculated (and implicitly the minimum number of preferences supported by ballots).

## Contents